\documentclass[10pt,journal,compsoc]{IEEEtran}
\ifCLASSOPTIONcompsoc
\usepackage[nocompress]{cite}
\else
\usepackage{cite}
\fi
\ifCLASSINFOpdf
\usepackage[pdftex]{graphicx}
\usepackage{multirow}
\usepackage{amsmath}
\usepackage{amssymb}
\usepackage{booktabs}
\usepackage{caption}

\usepackage{xcolor} 
\usepackage{color} 
\usepackage{verbatim} 
\usepackage{diagbox}
\usepackage[pagebackref=true,breaklinks=true,letterpaper=true,colorlinks,citecolor=blue,linkcolor=blue,bookmarks=false]{hyperref}
\else
\fi
\usepackage{graphicx}  
\usepackage{float}
\usepackage{subfigure} 

\begin{document}
\title{HHAvatar: Gaussian Head Avatar\\ with Dynamic Hairs}

\author{
Zhanfeng Liao\rm{*}, 
Yuelang Xu\rm{*}, 
Zhe Li,
Qijing Li, 
Boyao Zhou, 
Ruifeng Bai, 
Di Xu, \\
Hongwen Zhang, 
Yebin Liu
\thanks{
\indent\rm{*} indicates equal contribution.\\
\indent Zhanfeng Liao, Yuelang Xu, Zhe Li, Qijing Li, Boyao Zhou, and Yebin Liu are with the Department of Automation, Tsinghua University, Beijing 100084, P.R.China.\\
\indent Ruifeng Bai and Di Xu are with Huawei Technologies Co Ltd, Huawei Cloud, Shenzhen, Guangdong, P.R.China.\\
\indent Hongwen Zhang is with the School of Artificial Intelligence, Beijing Normal University, Beijing, P.R.China.\\
\indent Corresponding author: Yebin Liu.
}
    }

\newcommand{\mname}{ACLRNet}
\markboth{subimit to IEEE Transactions on Pattern Analysis and Machine Intelligence,~Vol.~XX, No.~XX, XX~2024}%
{Shell \MakeLowercase{\textit{Zamir et al.}}: Bare Demo of IEEEtran.cls for Computer Society Journals}

\IEEEtitleabstractindextext{
\begin{abstract}
Creating high-fidelity 3D head avatars has always been a research hotspot, but it remains a great challenge under lightweight sparse view setups.
%
In this paper, we propose HHAvatar represented by controllable 3D Gaussians for high-fidelity head avatar with dynamic hair modeling.
We first use 3D Gaussians to represent the appearance of the head, and then jointly optimize neutral 3D Gaussians and a fully learned MLP-based deformation field to capture complex expressions. 
The two parts benefit each other, thereby our method can model fine-grained dynamic details while ensuring expression accuracy. 
Furthermore, we devise a well-designed geometry-guided initialization strategy based on implicit SDF and Deep Marching Tetrahedra for the stability and convergence of the training procedure.
To address the problem of dynamic hair modeling, we introduce a hybrid head model into our avatar representation based Gaussian Head Avatar and a training method that considers timing information and an occlusion perception module to model the non-rigid motion of hair. 
Experiments show that our approach outperforms other state-of-the-art sparse-view methods, achieving ultra high-fidelity rendering quality at 2K resolution even under exaggerated expressions and driving hairs reasonably with the motion of the head. Project page: \url{https://liaozhanfeng.github.io/HHAvatar}.
\end{abstract}
    
}
\maketitle
\IEEEdisplaynontitleabstractindextext

\IEEEpeerreviewmaketitle

\IEEEraisesectionheading{\section{Introduction}\label{sec:introduction}}
\IEEEPARstart{H}{igh-fidelity} 3D human head avatar modeling is of great significance in many fields, such as VR/AR, telepresence, digital human and film production. 
Although some traditional head avatars \cite{lombardi2018deep, lombardi2021mixture, ma2021pixel, wang2023neural} realize high-fidelity animation, they typically require accurate geometries reconstructed and tracked from dense multi-view videos, thus limiting their applications in lightweight settings.
On the other hand, recent works~\cite{zhao2023havatar, raj2020pva, mihajlovic2022keypointnerf} have verified that Neural Radiance Fields (NeRF)~\cite{mildenhall2020nerf} can skip the geometry reconstruction and the tracking steps but directly learn high-quality NeRF-based head avatars in dense or sparse views, greatly lowering the threshold for head avatar reconstruction. 
However, it still remains challenging for these NeRF-based approaches to synthesize high-fidelity images at 2K resolutions with pixel-level details, including hairs, wrinkles, and eyes.

Recently, 3D Gaussian Splatting (3DGS)~\cite{kerbl3Dgaussians}, an explicit and efficient point-based representation, has been proposed for both high-fidelity rendering quality and real-time rendering speed.
Compared to NeRF, the reconstruction quality of static and dynamic scenes~\cite{wu20234d, luiten2023dynamic, yang2023realtime} is much better while rendering time cost has been significantly reduced. 
Some works~\cite{qian2024gaussianavatars,shao2024splattingavatar,zhao2024psavatar,chen2024monogaussianavatar,rivero2024rig3dgs,xiang2024flashavatar,wang2023gaussianhead} have verified 3DGS can also create photorealistic head avatars that are controllable in terms of expression and pose. 
Nevertheless, these methods continue to face challenges in generating high-fidelity images at 2K resolutions with precise pixel-level details, faithfully representing highly complex and exaggerated facial expressions, and capturing dynamic hairs.
To overcome this bottleneck and further improve the avatar quality, we propose Gaussian Head Avatar with Dynamic Hairs (HHAvatar), a novel representation that utilizes 3DGS for ultra high-fidelity head avatar with dynamic hair modeling. 

Previous explicit~\cite{zheng2023pointavatar} and implicit~\cite{athar2022rignerf,zheng2022imavatar, zielonka2022instant} head avatars usually formulate the facial deformation via linear blend skinning (LBS) using the skinning weights and blendshapes like the FLAME model~\cite{li2017learning}.
However, such a LBS-based formulation fails to represent exaggerated and fine-grained expressions by simple linear operations, limiting the representation ability of the head avatars.
Inspired by NeRSemble~\cite{kirschstein2023nersemble}, we propose a fully learnable expression-conditioned deformation field for the 3D head Gaussians, avoiding the limited capability of the LBS-based formulation.
Specifically, we input the positions of the 3D Gaussians with expression coefficients into a MLP to directly predict the displacements from the neutral expression to the target one.
Similarly, we control the motion of non-face areas, such as the neck, using the head pose as the condition.
3D Gaussian-based representation has the powerful ability to reconstruct high-frequency details, enabling our method to learn accurate deformation fields. 
In turn, the learned accurate deformation field facilitates the dynamic Gaussian head model to fit more dynamic details.
As a result, our method is able to reconstruct finer-grained dynamic details of expressive human heads.

\begin{figure*}[ht]
  \centering
  \includegraphics[width=\linewidth]{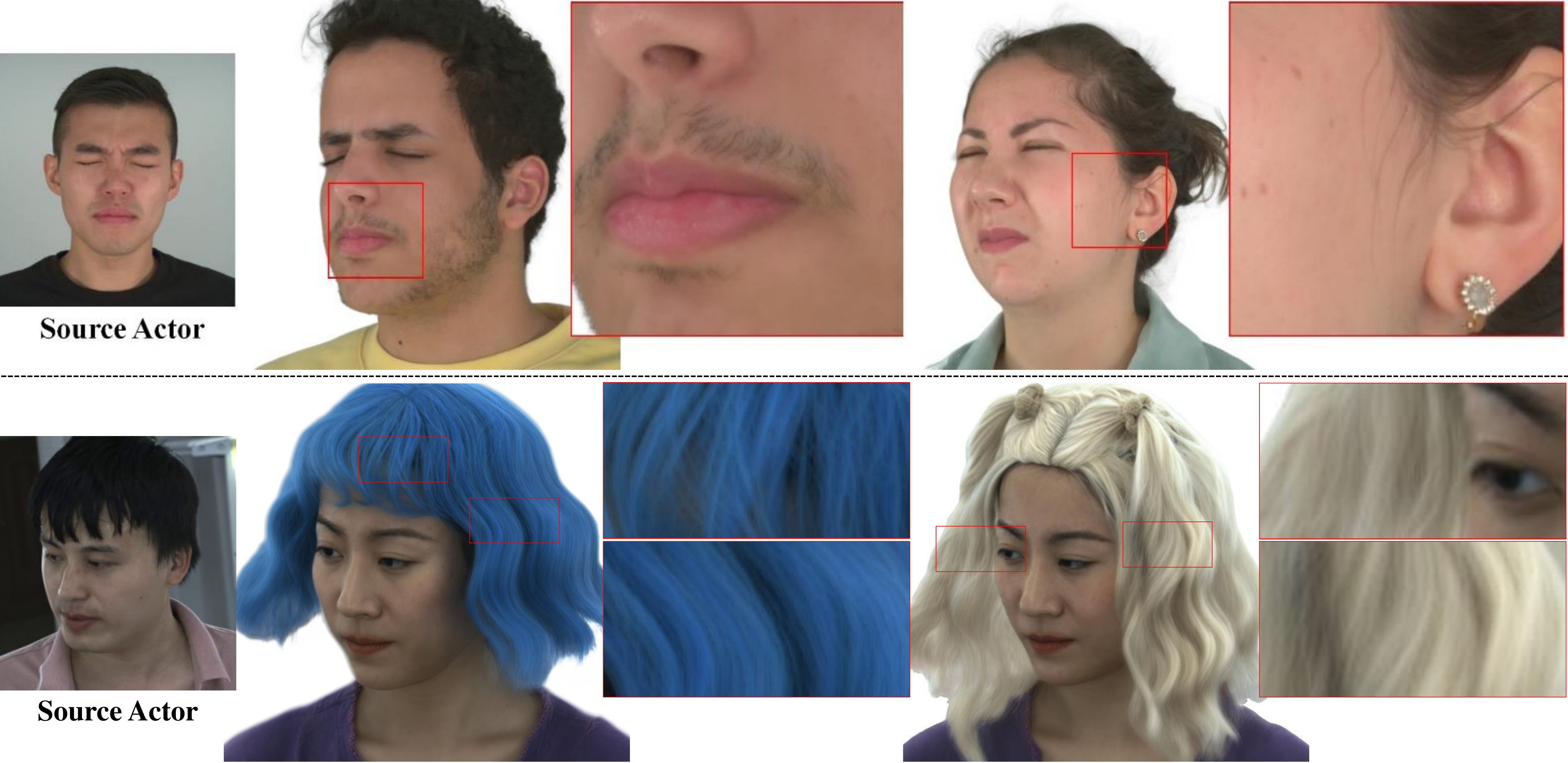}
  \caption{HHAvatar achieves ultra high-fidelity image synthesis with controllable expressions at 2K resolution. The above shows different identities animated by the same expression. 
  The bottom shows that variations in hair positions can arise for identical poses, stemming from diverse hair status (i.e., position and speed) at the previous moment. 
  }
  \label{fig:teaser}
\end{figure*}
Unfortunately, even though using MLPs can fit more dynamic details, it is not easy to use the same method to model the physical motion of hairs.
%
%
The reason is that the face can be determined by facial expressions and pose, but hair also needs to consider the speed and previous position due to inertia.
Meanwhile, some works~\cite{sklyarova2023neural,rosu2022neural,wang2023neuwigs,wang2022hvh,luo2024gaussianhair} mainly focus on hair reconstruction, but can not animate the head and the hair by facial expressions and head pose (i.e., not avatars).
%
To this end, we propose to use more suitable representations for different head components to develop head avatars capable of simulating the physical motion of hair.
%
%
Specifically, for hair modeling, we first obtain the 3DGS representation of canonical space, and then use MLP which takes hair status (i.e., position and speed) at the previous moment and the motion of the head as additional condition inputs, similar to deformable 3DGS~\cite{yang2024deformable}, to deform and capture hair dynamics.
Meanwhile, by employing mask supervision that introduces facial and hair occlusion, we segment the hair from the head and maintain the completeness of the head.
Furthermore, we introduce a temporal module for fusing time series information as the motion of hair is influenced by the speed and position of the previous moment.
%
%
As a result, our model can produce realistic animation with physical motion of the hairs.
%

As a discretized representation, the gradients back-propagated to the 3D Gaussians cannot spread through the whole space. 
Thus the convergence of training heavily relies on a plausible initialization for both the geometry and the deformation field.
However, simply initializing the 3D head Gaussians with a morphable template like FLAME~\cite{li2017learning} fails to model the long hairstyle and the shoulders.
Hence, we further propose an efficient and well-designed geometry-guided initialization strategy. 
This strategy not only initializes the parts included in the morphable template, but also provides a better initialization for dynamic hairs.
Specifically, instead of starting from stochastic Gaussians or a FLAME model, we initially optimize an implicit signed distance function (SDF) field along with a color field and a deformation MLP for modeling the basic geometry, color, and the expression-conditioned deformations of the head avatar respectively.
The SDF field is converted to a mesh through Deep Marching Tetrahedra (DMTet)~\cite{shen2021dmtet}, with the color and deformation of the vertices predicted by the MLPs.
Then we render the mesh and optimize them jointly under the supervision of multi-view RGB images.
Finally, we use the mesh with per-vertex features from the SDF field to initialize the 3D Gaussians to lie on the basic head surface while the color and deformation MLPs are carried over to the next stage, ensuring stable training for convergence. The entire initialization process takes only around 10 minutes.


%
A preliminary version of this work has been published in CVPR 2024~\cite{xu2023gaussianheadavatar}, in which we propose a novel head avatar representation for modeling realistic animatable head avatars under lightweight sparse-view setups. 
In this paper, we extend it to handle avatar modeling of dynamic hairs.
%
The contributions of our method can be summarized as:
\begin{itemize}
\item We propose HHAvatar, a new head avatar representation that employs controllable dynamic 3D Gaussians to model expressive human head avatars, producing ultra high-fidelity synthesized images at 2K resolutions. For modeling high-frequency dynamic details, we employ a fully learned deformation field upon the 3D head Gaussians, which accurately models extremely complex and exaggerated facial expressions and dynamic hair motion.

\item As far as we know, we propose the first head avatar that can model the physical motion of the hairs. 
By modeling hair separately from the head and introducing the kinematic features of the hair, the hair movement can be realistically simulated based on the motion of the head. By designing a training method that considers timing information and an occlusion perception module, dynamic hairs can be reconstructed under lightweight sparse-view setups.





\item We design an efficient initialization strategy that leverages implicit representations to initialize the geometry and deformation, leading to efficient and robust convergence when training the HHAvatar. This strategy not only initializes the parts included in the morphable template, but also provides a better initialization for dynamic hairs.

\end{itemize}
Benefiting from these contributions, our method surpasses recent state-of-the-art methods under lightweight sparse-view setups on the avatar quality by a large margin and model the physical motion of the hairs as shown in
Fig.~\ref{fig:teaser}.

\section{Related Work}
%
In this section, we will briefly present the research related to head avatars and data-driven hair animation.
\subsection{3D Head Avatar Modeling}
Due to the wide application value in the film and digital human industry, 3D head avatar reconstruction from multi-view images has always been a research hotspot. 
However, most of the works only focus on reconstruction without hair motion, and even only facial reconstruction.
Traditional works~\cite{levoy2000the, beeler2010high, ghosh2011multiview, bradley2010high} reconstruct the scan geometry through multi-view stereo and then register a face mesh template to it. 
However, such methods usually require heavy computation. With the utilization of deep neural networks, current methods~\cite{li2021topologically, timo2023instant, xiao2022detailed, yang2023asm} achieve very fast reconstruction, producing even more accurate geometry. 
Lombardi et al.~\cite{lombardi2018deep}, Bi et al.~\cite{bi2021deep} and Ma et al.~\cite{ma2021pixel} represent the full head mesh through a deep neural network and train it with multi-view videos as supervision. However, due to the errors in geometric estimation, mesh-based head avatars typically suffer from texture blur.
Therefore, some recent methods~\cite{wang2021learning, lombardi2019neural} utilize NeRF representation~\cite{mildenhall2020nerf} to synthesize novel view images without geometry reconstruction, or build NeRF on the head mesh template~\cite{lombardi2021mixture}. 
Furthermore, the NeRF-based methods are extended to sparse view reconstruction tasks~\cite{zhao2023havatar, mihajlovic2022keypointnerf, raj2020pva, kirschstein2023nersemble} and achieve impressive performance.

Methods which focus on generative model~\cite{paysan2009a, li2017learning, cao2014facewarehouse, brunton2014multilinear, blanz1999morphable, wu2023ganhead, wang2022faceverse} are dedicated to learning general mesh face templates from large-scale multi-view face images or 3D scans. Recently, implicit SDF-based~\cite{yenamandra2020i3dmm} or NeRF-based~\cite{zhuang2022mofanerf, hong2022headnerf, wang2022morf, cao2022authentic, sun2023next3d} methods can learn full-head templates without the limitations of fixed topology, thereby better modeling complex hairstyles and glasses. Cao et al.~\cite{cao2022authentic} adopts a hybrid representation of local NeRF built on the mesh surface, which enables high-fidelity rendering and flexible expression control.

3D head avatars reconstruction from monocular videos is also a popular yet challenging research topic. Early methods~\cite{cao2015real, cao2016real, ichim2015dynamic, hu2017avatar, deng2019accurate, nagano2018pagan} optimize a morphable mesh to fit the training video. Recent methods~\cite{grassal2022neural, Khakhulin2022realistic} leverage neural networks to learn non-rigid deformation upon 3DMM face templates~\cite{li2017learning, gerig2018morphable}, thus can recover more dynamic details. Such methods are not flexible enough to handle complex topologies. Therefore, the latest methods explore to construct head avatar models based on implicit SDF~\cite{zheng2022imavatar}, point clouds~\cite{zheng2023pointavatar} or NeRF~\cite{guo2021ad, liu2022semantic, gafni2021dynamic, athar2021flame, athar2022rignerf, gao2022reconstructing, xu2023avatarmav, zielonka2022instant, xu2023latentavatar, qin2023high}.

\subsection{Head Avatar by Gaussian Splatting}
Point elements as a discrete and unstructured representation can fit geometry with arbitrary topology~\cite{yifan2019differentiable} efficiently. Recent methods~\cite{wiles2020synsin, lassner2021pulsar, kopanas2022neural} open up a differentiable rasterization pipeline, such that the point-based representation is widely used in multi-view reconstruction tasks. Aliev et al.~\cite{aliev2020neural} and Ruckert et al.~\cite{ruckert2022adop} propose to first render the feature map, which is transferred to the images through a convolutional renderer. Xu et al.\cite{xu2022point} use neural point cloud associated with neural features to model a NeRF. 

Recently, 3D Gaussian splatting~\cite{kerbl3Dgaussians} shows its superior performance, beating NeRF in both novel view synthesis quality and rendering speed. 
Some approaches~\cite{wu20234d, yang2023realtime, luiten2023dynamic, yang2023deformable, li2024animatablegaussians, zheng2023gpsgaussian, zheng2023gpsgaussian, shao2023control4d} extend Gaussian representation to dynamic scene reconstruction. 
However, these methods can not be migrated to the head avatar reconstruction tasks.
GaussianAvatars~\cite{qian2024gaussianavatars} create photorealistic head avatars that are fully controllable in terms of expression and pose from multi-view videos. 
While other approaches like SplattingAvatar~\cite{shao2024splattingavatar}, PSAvatar~\cite{zhao2024psavatar}, MonoGaussianAvatar~\cite{chen2024monogaussianavatar}, Rig3DGS~\cite{rivero2024rig3dgs}, FlashAvatar~\cite{xiang2024flashavatar} and GaussianHead~\cite{wang2023gaussianhead} employ the coefficients of 3DMM to control the head and reconstruct a 3DGS-based animatable head avatar from monocular videos. 
However, these approaches need the hair maintain as still as possible and model the hair as a part rigidly attached to the head.
%

\subsection{Data-driven Hair Animation}
In academia and the film and gaming industries, using physics based simulations to create hair animations is a common practice~\cite{bertails2008realistic}.
However, using physics based simulations to generate hair animations may be computationally expensive. 
To address this issue, a simplified data-driven approach~\cite{chai2014reduced,chai2016adaptive,guan2012multi} simulates only a small portion of guided hair bundles and interpolates the remaining parts using skin weights learned from the complete simulation.

With the latest advances in deep learning, the use of neural networks has improved the efficiency of dynamic generation~\cite{lyu2020real} and rendering~\cite{chai2020neural,olszewski2020intuitive} of hair.
Some methods~\cite{lyu2020real} use deep neural networks for adaptive binding between normal hair and guided hair. 
Some methods~\cite{olszewski2020intuitive} treat hair rendering as an image translation problem and generate realistic rendering of hair based on 2D hair masks and strokes. 
Some methods~\cite{chai2020neural} achieve faster rendering and realistic results by using screen space neural rendering technology instead of the rendering part in the animation pipeline.
However, these methods are still based on traditional hair simulation pipelines and use synthetic wigs, which require artists to manually set and are not easy to measure and evaluate.
Some methods~\cite{wu2016data} propose to use a secondary motion graph for hair animation at runtime, without relying on traditional hair simulation pipelines. 
However, this method is limited by the artist's design of wigs and control over hair simulation parameters, and cannot be used to animate real hair motions. 

Additionally, certain techniques require only multi-view videos and camera parameters as input without necessitating the manual design of wigs by artists~\cite{wang2023neuwigs,luo2024gaussianhair,zakharov2024human}. 
Some methods~\cite{wang2023neuwigs,luo2024gaussianhair} do not need the artist's design of wigs.
NeuWigs~\cite{wang2023neuwigs} uses the mixture of volumetric primitives (MVP) to model hair and does not rely on a manual hair design. 
However, this method needs a lot of cameras to capture a multi-view video for reconstruction, and only considers the driving of the hair and cannot complete the driving of the entire head. 
GaussianHair~\cite{luo2024gaussianhair} and Gaussian Haircut~\cite{zakharov2024human} use specially designed Gaussian representation to model hair, which can achieve high-quality hair reconstruction. However, this method requires first relying on multi-view images to complete static reconstruction, and then driving based on a conventional CG rendering engine, which is not data-driven. Moreover, this method cannot animate the head by facial expressions and head poses.
Compared with previous methods, our approach can achieve data-driven head modeling, including the face and the hair, under lightweight sparse-view setups.

\section{Method}
\label{sec:method}

\begin{figure*}[ht]
  \centering
  \includegraphics[width=\linewidth]{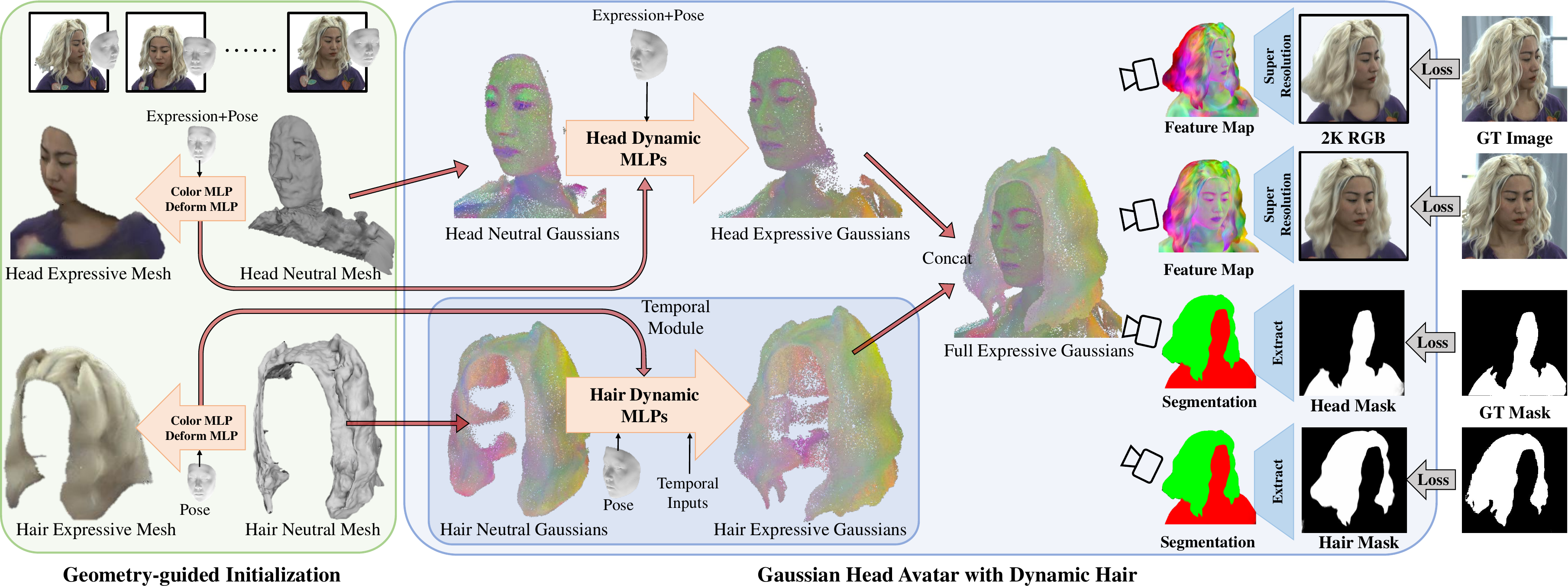}
  \caption{The pipeline of the HHAvatar rendering and reconstruction. We first optimize the guidance model including a neutral mesh, a deformation MLP and a color MLP in the Initialization stage. Then we use them to initialize the neutral Gaussians and the dynamic generator. Finally, 2K RGB images are synthesized through differentiable rendering and the super-resolution network, and the segmentation maps of the hair and the head are also synthesized through differentiable rendering. The HHAvatar are trained under the supervision of multi-view RGB videos and multi-view masks from face-parsing.}
  \label{fig:overview}
\end{figure*}

\begin{figure}[ht]
  \centering
  \includegraphics[width=\linewidth]{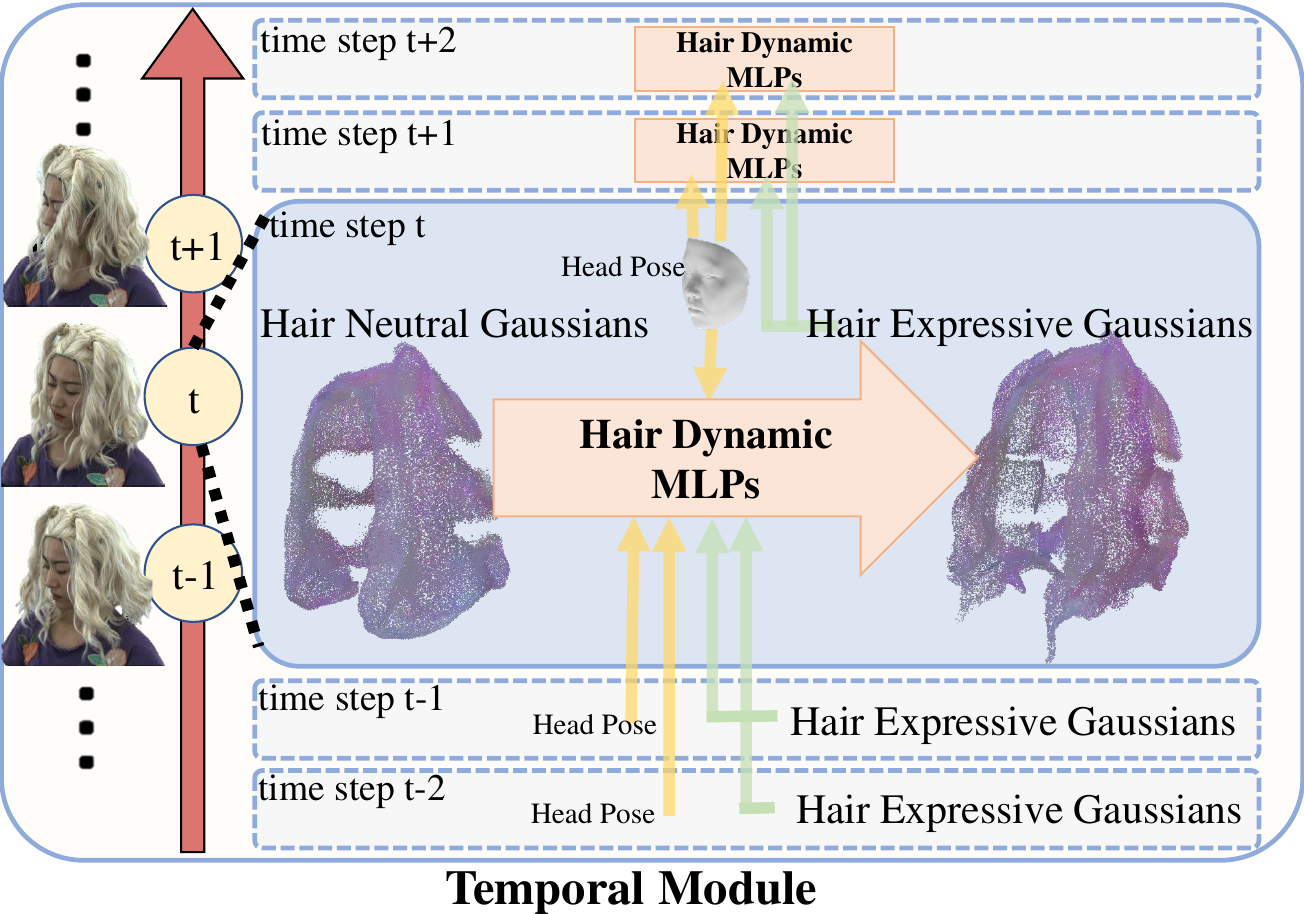}
  \caption{The detail of the temporal module. The input for the Hair Dynamic MLPs at time step $t$ is $\boldsymbol{X}_0$ (the position of the neutral Gaussian point), $\{X'_{t-1}, X'_{t-2}\}$ (the position of the expressive Gaussian point at time step $t-1$ and time step $t-2$), and $\{\beta_{t}, \beta_{t-1}, \beta_{t-2}\}$ (the pose of the head at time step $t$, $t-1$, and $t-2$). The specific details of the Hair Dynamic MLPs are detailed in Sec. \ref{subsubsec:dynamichair}.}
  \label{fig:overview_detail}
\end{figure}

The pipeline of the reconstruction of HHAvatar is illustrated in Fig.~\ref{fig:overview}, including the initialization stage and the training stage of HHAvatar.
The detail of the temporal module is illustrated in Fig.~\ref{fig:overview_detail}.
Before the beginning of the pipeline, we remove the background~\cite{Lin2021real} of each image and jointly estimate the 3DMM model~\cite{gerig2018morphable}, 3D facial landmarks and the expression coefficients for each frame, and we obtain the mask of the hair and the head by face-parsing~\cite{yu2021bisenet}~\footnote{https://github.com/zllrunning/face-parsing.PyTorch}.
The whole model is optimized under the supervision of multi-view RGB videos and hair masks.

%

\subsection{Avatar Representation}
\label{subsec:representation}

Generally, the static 3D Gaussians~\cite{kerbl3Dgaussians} with $N$ points are represented by their positions $X$, the multi-channel color $C$, the rotation $Q$, scale $S$ and opacity $A$. The rotation $Q$ is represented in the form of quaternion. Subsequently, the Gaussians can be rasterized and rendered to the multi-channel image $I$ given the camera parameters $\mu$. This process can be formulated as:
\begin{equation}
\label{eqn:render}
 I = \mathcal{R}(X, C, Q, S, A; \mu).
\end{equation}

Our task is to reconstruct a dynamic head avatar controlled by facial expression, head pose, and head speed. 
For the region outside the hair on the head, the head avatar only depends on the facial expression and the head pose.
For the hairs, a point $P_t$ in hairs at time step $t$ depend on the point $P_{t-1}$ at time step $t-1$, the speed of the point, the head pose at time step $t$, and the head speed. 
Therefore, we formulate the head avatar as dynamic 3D Gaussians conditioned on the facial expression, the head pose, head speed, and the points in hairs at the previous 2 time steps. 
To handle the dynamic changes, we input the above conditions to the head avatar model and output the position and other attributes of the Gaussians as above.

Specifically, we first extract the neutral mesh for the head and the hair through DMTet~\cite{shen2021dmtet} to initialize the neutral Gaussians (Sec.~\ref{subsec:initialization}) and construct a canonical neutral Gaussian model with expression-independent attributes including $\boldsymbol{X}_0 \in \mathbb{R}^{N \times 3}$ denotes the positions of the Gaussians with a neutral expression in the canonical space, $\boldsymbol{F}_0 \in \mathbb{R}^{N \times 128}$ denotes the point-wise feature vectors as their intrinsic properties for generating high-resolution images, $\boldsymbol{Q}_0 \in \mathbb{R}^{N \times 4}$, $\boldsymbol{S}_0 \in \mathbb{R}^{N \times 3}$, and $\boldsymbol{A}_0 \in \mathbb{R}^{N \times 1}$ denote the neutral rotation, scale and opacity respectively. 
$\boldsymbol{X}_0$, $\boldsymbol{F}_0$, $\boldsymbol{Q}_0$, $\boldsymbol{S}_0$, and $\boldsymbol{A}_0$ are fully optimizable. 
Note that we do not define the neutral color $\boldsymbol{C}_0$, but directly predict expression-dependent dynamic color from the point-wise feature vectors $\boldsymbol{F}_0$.
Then, we construct a MLP-based expression conditioned dynamic generator $\boldsymbol{\Phi}$ to generate all the extra dynamic changes to the neutral model.
Because the deformation of hair at time step $t$ is related to the hair position $X_{t-1}$ at time step $t-1$, the hair speed which can get from $X_{t-1}-X_{t-2}$, the head pose $\beta_{t}$ at time step $t$, and the head movement speed which can get from $\beta_{t}$ and $\beta_{t-1}$, the dynamic generator $\boldsymbol{\Phi}$ needs $X_{t-1}$, $X_{t-2}$, and $\beta_{t-1}$ as extra inputs.
Overall, the whole HHAvatar can be formulated as:
\begin{equation}
\label{eqn:whole_model}
\begin{split}
 \{X_{t}, C_{t}, Q_{t}, S_{t}, A_{t}\} = \boldsymbol{\Phi}(\boldsymbol{X}_0, \boldsymbol{F}_0, \boldsymbol{Q}_0, \boldsymbol{S}_0, \boldsymbol{A}_0; \\ X_{t-1}, X_{t-2}, \theta_{t}, \beta_{t}, \beta_{t-1}),
 \end{split}
\end{equation}
with $\{X_{t}, C_{t}, Q_{t}, S_{t}, A_{t}\}$ denoting $\{X, C, Q, S, A\}$ at the time step $t$ and $\theta_{t}$ denoting expression coefficients at the time step $t$.
During the training, we optimize all the parameters of the dynamic generator $\boldsymbol{\Phi}$ and the neutral Gaussian model $\{\boldsymbol{X}_0, \boldsymbol{F}_0, \boldsymbol{Q}_0, \boldsymbol{S}_0, \boldsymbol{A}_0\}$, which are highlighted in bold in the following.

\subsubsection{Head Avatar}
Next, we explain the process of adding expression-related changes to the neutral Gaussian model through the dynamic generator $\boldsymbol{\Phi}$ as described in Eqn.~\ref{eqn:whole_model} in detail.
We use superscript (') to denote the parameters of Gaussians which are changed by MLPs in the canonical space but not transformed to the world space by head pose $\beta_t$.

\textbf{Positions} $X'_{t}$ of the Gaussians at time step $t$. Expressions bring about the geometric deformation of the neutral model, which is modeled as the displacements of the Gaussian points. Specifically, we predict the displacements respectively controlled by the expression and the head pose in the canonical space through two different MLPs: $\boldsymbol{f}_{def}^{exp} \in \boldsymbol{\Phi}$ and $\boldsymbol{f}_{def}^{pose} \in \boldsymbol{\Phi}$. Then, we add them to the neutral positions. 
\begin{equation}
\label{eqn:deform}
\begin{split}
 X'_{t} = \boldsymbol{X}_0 + \lambda_{exp}(\boldsymbol{X}_0) \boldsymbol{f}_{def}^{exp}(\boldsymbol{X}_0, \theta_{t}) \\
 + \lambda_{pose}(\boldsymbol{X}_0) \boldsymbol{f}_{def}^{pose}(\boldsymbol{X}_0, \beta_{t}).
\end{split}
\end{equation}
$\lambda_{exp}(\cdot)$ and $\lambda_{pose}(\cdot)$ represent the extent to which the points are affected by the expression or the head pose respectively.
Without decoupling but just using the expression coefficients as the global condition~\cite{gafni2021dynamic} which also controls the shoulders and upper body, will produce jittering results during the animation.
Here, we assume that the Gaussian points closer to 3D landmarks are more affected by the expression coefficients and less affected by the head pose, while the opposite is true for the Gaussian points far away. Specifically, The 3D landmarks $\boldsymbol{P}_0$ of the canonical model are first estimated through the 3DMM model in the data preprocessing and then optimized in the initialization stage~\ref{subsec:initialization}. Then for each Gaussian point, we calculate the above weight $\lambda_{exp}(\cdot)$ and $\lambda_{pose}(\cdot)$ as follows:
\begin{align*}
\begin{split}
\lambda_{exp}(x)= \left \{
\begin{array}{ll}
    1,                                                               & dist(x, \boldsymbol{P}_0) < t_{1} \\
    \frac{t_{2}-dist(x, \boldsymbol{P}_0)}{t_{2}-t_{1}},               & dist(x, \boldsymbol{P}_0) \in [t_{1},t_{2}]\\
    0,                                                               & dist(x, \boldsymbol{P}_0) > t_{2},
\end{array}
\right.
\end{split}
\end{align*}
with $\lambda_{pose}(x) = 1 - \lambda_{exp}(x)$. And $x \in \mathbf{X}_0$ denotes the position of one neutral Gaussian. 
$dist(x, \boldsymbol{P}_0)$ denotes the minimum distance from the point $x$ to the 3D landmarks $\boldsymbol{P}_0$. $t_{1}=0.15$ and $t_{2}=0.25$ are predefined hyperparameters when the length of the head is set to approximately $1$. For the hairs, we set $\lambda_{exp}(x) = 0$ since the hairs are hardly affected by the
expression coefficients.

\textbf{Color} $C'_{t}$ of the Gaussians at time step $t$. Modeling the dynamic details typically requires dynamic color that changes with expressions. As we do not pre-define the neutral value in Eqn.~\ref{eqn:whole_model}, the color are directly predict by two color MLPs: $\boldsymbol{f}_{col}^{exp} \in \boldsymbol{\Phi}$ and $\boldsymbol{f}_{col}^{pose} \in \boldsymbol{\Phi}$: 
\begin{equation}
\label{eqn:color}
\begin{split}
 C'_{t} = \lambda_{exp}(\boldsymbol{X}_0) \boldsymbol{f}_{col}^{exp}(\boldsymbol{F}_0, \theta_{t}) \\
 + \lambda_{pose}(\boldsymbol{X}_0) \boldsymbol{f}_{col}^{pose}(\boldsymbol{F}_0, \beta_{t}).
\end{split}
\end{equation}

\textbf{Rotation}, \textbf{Scale} and \textbf{Opacity} $\{Q'_{t}, S'_{t}, A'_{t}\}$ of the Gaussians at time step $t$. These three attributes also dynamic, thereby modeling some detailed expressions-related appearance changes. Here, we just use another two attribute MLPs $\boldsymbol{f}_{att}^{exp} \in \boldsymbol{\Phi}$ and $\boldsymbol{f}_{att}^{pose} \in \boldsymbol{\Phi}$ to predict their shift from the neutral value.
\begin{equation}
\begin{split}
 \{Q'_{t}, S'_{t}, A'_{t}\} = \{\mathbf{Q}_0, \mathbf{S}_0, \mathbf{A}_0\} \\
 + \lambda_{exp}(\boldsymbol{X}_0) \boldsymbol{f}_{att}^{exp}(\boldsymbol{F}_0, \theta_{t}) \\
 + \lambda_{pose}(\boldsymbol{X}_0) \boldsymbol{f}_{att}^{pose}(\boldsymbol{F}_0, \beta_{t}).
\end{split}
\end{equation}

\subsubsection{Dynamic Hair}
\label{subsubsec:dynamichair}
Next, we explain the process of the changes in hair with head motion in detail.
We assume the head and the hair maintain still at time step $0$, which means that hairs do not have extra deformation due to the inertia, the hair positions $X_0=X_{-1}=X_{-2}$, and the head poses $\beta_0=\beta_{-1}=\beta_{-2}$.

\textbf{Hair deformation.} The deformation of hair is related to the previous hair position $X_{t-1}$ which can get from $X'_{t-1}$ and $\beta_{t-1}$, hair speed which can get from $X'_{t-1}$, $X'_{t-2}$, $\beta_{t-1}$, and $\beta_{t-2}$, head pose $\beta_t$, and head movement speed which can get from $\beta_{t}$ and $\beta_{t-1}$. 
We use a MLP $\boldsymbol{f}_{def}^{hair} \in \boldsymbol{\Phi}$ to predict the extra hair deformation.

\begin{equation}
\label{eqn:deform_hair}
\begin{split}
 \resizebox{.89\hsize}{!}{$
 X'_t = \boldsymbol{X}_0 + \lambda_{hair}(\boldsymbol{X}_0) \boldsymbol{f}_{def}^{hair}(\boldsymbol{X}_0, X'_{t-1, t-2}, \beta_{t, t-1, t-2})$}.
\end{split}
\end{equation}
Here, $X'_{t-1, t-2}$ and $\beta_{t, t-1, t-2}$ represent respectively the $X'$ at time step ${t-1, t-2}$ and the $\beta$ at time step ${t, t-1, t-2}$. $\lambda_{hair}(\cdot)$ represents the extent to which the points of the hair are affected by the head pose. 
The strength of the hair deformation is related to the distance between the hairs and the scalp.
We assume that the Gaussian points closer to the scalp are more affected by the head pose, while the opposite is true for the Gaussian points far away. Specifically, The scalp $\boldsymbol{P}_1$ of the canonical model are first estimated through the FLAME model~\cite{li2017learning} in the data preprocessing. Then for each Gaussian point, we calculate the above weight $\lambda_{exp}(\cdot)$ and $\lambda_{pose}(\cdot)$ as follows:
\begin{align*}
\begin{split}
\lambda_{hair}(x)= \left \{
\begin{array}{ll}
    1,                                                               & dist(x, \boldsymbol{P}_1) < t_{3} \\
    \frac{t_{4}-dist(x, \boldsymbol{P}_1)}{t_{4}-t_{3}},               & dist(x, \boldsymbol{P}_1) \in [t_{3},t_{4}]\\
    0,                                                               & dist(x, \boldsymbol{P}_1) > t_{4},
\end{array}
\right.
\end{split}
\end{align*}
with $x \in \mathbf{X}_0$ denotes the position of one neutral Gaussian. 
$dist(x, \boldsymbol{P}_1)$ denotes the minimum distance from the point $x$ to the scalp $\boldsymbol{P}_1$. $t_{3}=0.05$ and $t_{4}=0.15$ are predefined hyperparameters when the length of the head is set to approximately $1$. 

\textbf{Hair attribution.} The Gaussian properties of hair should also undergo corresponding changes after deformation. Here we use two MLPs $\boldsymbol{f}_{col}^{hair}$ and $\boldsymbol{f}_{att}^{hair}$ $\in \boldsymbol{\Phi}$ to predict the extra changes in the Gaussian properties of hair. 

\begin{equation}
\begin{split}
C'_{t} = C'_{t} + \boldsymbol{f}_{col}^{hair}(\boldsymbol{X}_0, X'_{t-1, t-2}, \beta_{t, t-1, t-2}),
\end{split}
\end{equation}

\begin{equation}
\begin{split}
 \resizebox{.89\hsize}{!}{$\{Q'_{t}, S'_{t}\} = \{\boldsymbol{Q}_0, \boldsymbol{S}_0\} 
 + \boldsymbol{f}_{att}^{hair}(\boldsymbol{X}_0, X'_{t-1, t-2}, \beta_{t, t-1, t-2})$}.
\end{split}
\end{equation}
Finally, we apply rigid rotations and translations $T(\cdot)$ to the Gaussians, transforming them from the canonical space to the world space. Note, the transformation is only implemented for directional variables: $\{X'_t, Q'_t\}$, while the multi-channel color, the scale and the opacity $\{C'_t, S'_t, A'_t\}$ are not directional thus remain unchanged.
\begin{equation}
 \{X_t, Q_t\} = T(\{X'_t, Q'_t\}, \beta_t),
\end{equation}
\begin{equation}
 \{C_t, S_t, A_t\} = \{C'_t, S'_t, A'_t\}.
\end{equation}


\subsection{Training}
\label{subsec:avatar_learning}
In this part, we explain the training pipeline of the HHAvatar~\ref{subsec:representation} and the loss function.
In each iteration, we first generate the expression conditioned 3D Gaussians as Eqn.~\ref{eqn:whole_model}. Then, given a camera view, we render a 32-channel image with 512 resolution $I_{C} \in \mathbb{R}^{512 \times 512 \times 32}$ referring to Eqn.~\ref{eqn:render}. After that we feed the image to a super resolution network $\boldsymbol{\Psi}$ to generate a 2048 resolution RGB image $I_{hr} \in \mathbb{R}^{2048 \times 2048 \times 3}$, such that more details are recovered and noise caused by uneven ambient light or camera chromatic aberration in the training data will be filtered out~\cite{xu2023latentavatar, ruckert2022adop}. Meanwhile, the expressive Gaussians are also rendered to a segmentation map for separating the hair from the head.

During training, we jointly optimize all the learnable parameters mentioned above in bold, including the neutral Gaussians: $\{\boldsymbol{X}_0, \boldsymbol{F}_0, \boldsymbol{Q}_0, \boldsymbol{S}_0, \boldsymbol{A}_0\}$, the dynamic generator: $\{\boldsymbol{f}_{col}^{exp}, \boldsymbol{f}_{col}^{pose}, \boldsymbol{f}_{def}^{exp}, \boldsymbol{f}_{def}^{pose}, \boldsymbol{f}_{att}^{exp}, \boldsymbol{f}_{att}^{pose}
\boldsymbol{f}_{def}^{hair},
\boldsymbol{f}_{col}^{hair},
\boldsymbol{f}_{att}^{hair}
\}$, and the super resolution network $\boldsymbol{\Psi}$. 
Our training loss consists of two parts: the reconstruction loss $\mathcal{L}_{recon}$ and the mask loss $\mathcal{L}_{mask}$:
\begin{equation}
\begin{split}
 \mathcal{L} = \mathcal{L}_{recon} + \lambda_{mask}\mathcal{L}_{mask}.
\end{split}
\end{equation}

\textbf{Reconstruction loss.} 
For the reconstruction loss function, we only use the foreground RGB images $I_{gt}$ as supervision to construct an L1 loss and a VGG perceptual loss~\cite{zhang2018the} between the generated images $I_{hr}$ and the ground truth $I_{gt}$. Besides, we encourage the first three channels of the 32-channel feature image $I_{C}$ to be RGB channels, which is ensured by a L1 loss term. 
The total loss is:
\begin{equation}
\begin{split}
 \mathcal{L}_{recon} = ||I_{hr} - I_{gt}||_{1} + \lambda_{vgg} VGG(I_{hr}, I_{gt}) \\
 + \lambda_{lr}(||I_{lr} - I_{gt}||_{1} + \lambda_{vgg} VGG(I_{lr}, I_{gt})),
\end{split}
\end{equation}
with $I_{lr}$ denoting the first three channels of the 32-channel image $I_{C}$. We set the weights $\lambda_{vgg}=0.1$ and $\lambda_{lr}=1$. 

\textbf{Mask loss.} To separate the points of the head and the hair, we introduce the occlusion perception module to use the head mask and the hair mask to supervise the head part and the hair part. We first use face-parsing~\cite{yu2021bisenet} to obtain the mask of the hair and the head. Then, we set the Gaussian color of the head and the hair to $(1.0, 0.0, 0.0)$ and $(0.0, 1.0, 0.0)$ respectively. Finally, we can get the predicted color by rendering to obtain the head mask and the hair mask, which can avoid supervision of the parts covered by hair.
\begin{equation}
\begin{split}
 \mathcal{L}_{mask} = ||M'_{head} - M_{head}||_{1} + ||M'_{hair} - M_{hair}||_{1},
\end{split}
\end{equation}
with $M'_{head}$ and $M'_{hair}$ denoting the predicted head mask and the predicted hair mask respectively, $M_{head}$ and $M_{hair}$ denoting the head mask and the hair mask from face-parsing~\cite{yu2021bisenet}. We set the weight $\lambda_{mask}=0.1$. 


\subsection{Geometry-guided Initialization}
\label{subsec:initialization}
Unlike neural networks, the Gaussians act as an unordered and unstructured representation. Random initialization leads to failure to converge while naively using the FLAME model to initialize will significantly reduce the reconstruction quality. In this section, we describe in detail how to optimize a mesh guidance model to provide reliable initialization for the Gaussians in Sec.~\ref{subsec:representation}.

\textbf{Mesh Guidance Model.} Specifically, we first construct two MLPs $\boldsymbol{f}_{sdf\_head}$ and $\boldsymbol{f}_{sdf\_hair}$ to represent two signed distance fields for the head and the hair respectively. 
In addition, this network will also output the corresponding feature vector of each point, which is used for predicting the point color. 
It can be formulated as:
\begin{equation}
 {s_{head}, \eta_{head}} = \boldsymbol{f}_{sdf\_head}(x),
\end{equation}
\begin{equation}
 {s_{hair}, \eta_{hair}} = \boldsymbol{f}_{sdf\_{hair}}(x),
\end{equation}
with $s_{head}$ and $s_{hair}$ denote the SDF value of the head and the hair respectively, $\eta_{head}$ and $\eta_{hair}$ denote the feature vector of the head and the hair respectively and $x$ denotes the point position. 
Then through DMTet~\cite{shen2021dmtet}, we can differentially extract the mesh with vertices $X$, per-vertex feature vectors $F$, and its faces for the head and the hair respectively, and we merge the mesh of the head and the hair to obtain the complete mesh of the whole head. 
We also predict the per-vertex 32-channel color as Eqn.~\ref{eqn:color} by the two color MLPs $\boldsymbol{f}_{col}^{exp}$ and $\boldsymbol{f}_{col}^{pose}$ for the head and the hair. 
In parallel, we construct the two deformation MLPs: $\boldsymbol{f}_{def}^{exp}$ and $\boldsymbol{f}_{def}^{pose}$ as described in Sec.~\ref{subsec:representation} to predict the displacements and add them to the vertex positions. 
This process is similar to Eqn~\ref{eqn:deform} above, with the Gaussian positions $\boldsymbol{X}_0$ replaced by the vertex positions $X$. 
Finally, we also apply rigid rotations and translations to the deformed mesh, transforming it to the world space and rendering the deformed mesh into an image $I$, a mask $M$, a head mask $M'_{head}$, and a hair mask $M'_{hair}$ through differentiable rasterization~\cite{munkberg2022extracting} according to the camera parameters $\mu$. 
Note that during the geometry-guided initialization, we temporarily ignore the real physical motion of the hair (i.e., we do not consider the additional motion of hair due to inertia).

\textbf{Loss Function and Training.} Next, we can first construct the RGB loss and the silhouette loss to train the guidance model:
\begin{equation}
 \mathcal{L}_{RGB} = ||I_{r,g,b} - I_{gt}||_{1},
\end{equation}
\begin{equation}
 \mathcal{L}_{sil} = IOU(M, M_{gt}),
\end{equation}
with $I_{gt}$ and $M_{gt}$ denote the ground truth RGB image and mask, respectively. $IOU(\cdot)$ denotes Intersection over Union metrics. Note that only the first three channels ${R,G,B}$ of the 32-channel image $I$ are supervised by the ground truth RGB images.

We also use the estimated 3D facial landmarks $P_{gt}$ to provide rough guidance for the expression deformation MLP. Specifically, we input the neutral 3D landmarks $\boldsymbol{P}_0$ into the expression deformation MLP to predict the expression conditioned landmarks $P$:
\begin{equation}
 P = \boldsymbol{P}_0 + \boldsymbol{f}_{def}^{exp}(\boldsymbol{P}_0, \theta).
\end{equation}
Then we construct the loss function with 3D facial landmarks $P_{gt}$ as the supervision:
\begin{equation}
 \mathcal{L}_{def} = ||P - P_{gt}||_{2}.
\end{equation}

Besides, we introduce three constraints: 
(1) a regularization term $\mathcal{L}_{offset}$ to punish all non-zero displacements to prevent the two deformation MLPs from learning a global constant offset~\cite{xu2023avatarmav}, 
(2) a regularization term $\mathcal{L}_{lmk}$ to limit the SDF value at the 3D landmarks to be close to zero, such that the landmarks are located on the surface of the mesh, 
(3) a Laplacian term $\mathcal{L}_{lap}$ for maintaining the extracted mesh smooth to a certain extent.
To obtain the signed distance field of the head and the hair respectively, we also use the head mask and the hair mask to supervise the head part and the hair part:
\begin{equation}
\begin{split}
 \mathcal{L}_{mask} = IOU((1 - M_o) \cdot M'_{head}, M_{head})\\ 
 + IOU(M_o \cdot M'_{hair}, M_{hair}),
\end{split}
\end{equation}
\begin{equation}
 M_o = D_{hair} < D_{head},
\end{equation}
with $M'_{head}$ and $M'_{hair}$ denoting the predicted head mask and the predicted hair mask respectively, $M_{head}$ and $M_{hair}$ denoting the head mask and the hair mask from face-parsing~\cite{yu2021bisenet} respectively, and $D_{head}$ and $D_{hair}$ denoting the predicted head depth and the predicted hair depth respectively.

Overall, the total loss function in the initialization stage is formulated as:
\begin{equation}
\begin{split}
 \mathcal{L} = \mathcal{L}_{RGB} + \lambda_{sil}\mathcal{L}_{sil} + \lambda_{def}\mathcal{L}_{def} \\+ 
 \lambda_{offset}\mathcal{L}_{offset} + \lambda_{lmk}\mathcal{L}_{lmk} \\+ \lambda_{lap}\mathcal{L}_{lap} +
 \lambda_{mask}\mathcal{L}_{mask},
\end{split}
\end{equation}
with $\lambda$ denoting the weights of each term, which are set as follows: $\lambda_{sil}=0.1$, $\lambda_{def}=1$, $\lambda_{offset}=0.01$, $\lambda_{lmk}=0.1$, $\lambda_{lap}=100$, and $\lambda_{mask}=0.1$. We jointly optimize the MLPs mentioned above with the neutral 3D landmarks $\boldsymbol{P}_0$ jointly until all MLPs are converged. 

\textbf{Parameters Transfer.} Finally, we use the roughly trained mesh guidance model to initialize the Gaussian model. Specifically, we extract the neutral mesh with vertices $X$ and per-vertex features $F$ through DMTet~\cite{shen2021dmtet} for the head and the hair respectively, and directly assign their values to the neutral positions $\boldsymbol{X}_0=X$ and the per-vertex feature vectors $\boldsymbol{F}_0=F$ of the neutral Gaussians respectively. 
Then,  we retain all the four optimized MLPs: $\{\boldsymbol{f}_{col}^{exp}, \boldsymbol{f}_{col}^{pose}, \boldsymbol{f}_{def}^{exp}, \boldsymbol{f}_{def}^{pose}\}$ for the Gaussian model. For the other neutral attributes: rotation, scale and opacity, we adopt the original initialization strategy in Gaussian Splatting~\cite{kerbl3Dgaussians}. And the parameters of the two attribute MLPs: $\{\boldsymbol{f}_{att}^{exp}, \boldsymbol{f}_{att}^{pose}, \boldsymbol{f}_{def}^{hair}, \boldsymbol{f}_{col}^{hair}, \boldsymbol{f}_{att}^{hair}\}$ and the super resolution network $\boldsymbol{\Psi}$ are just randomly initialized.

\section{Experiments}
\label{sec:experiments}

\begin{figure*}
  \centering
  \includegraphics[width=1.0\linewidth]{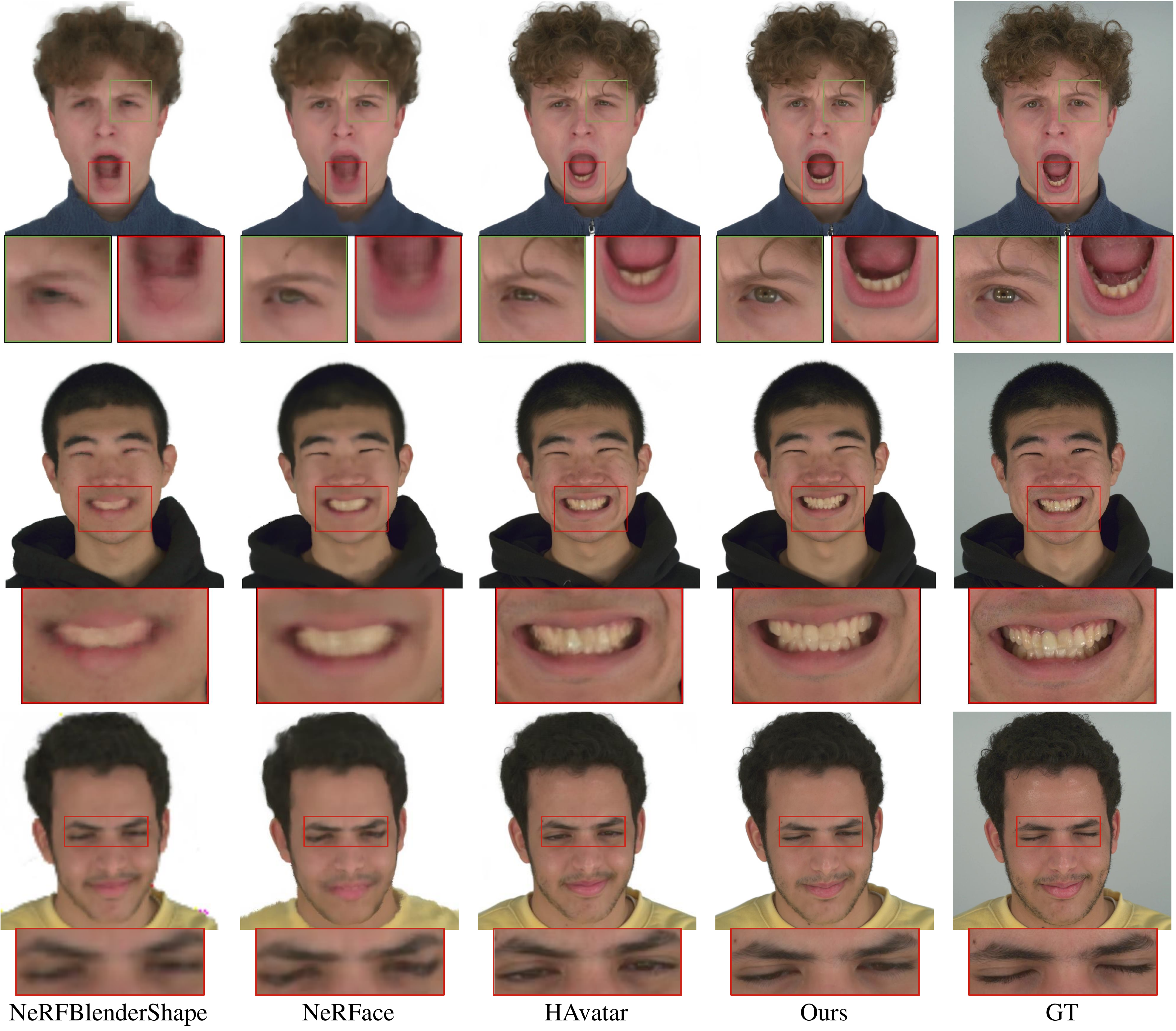}
  \caption{Qualitative comparisons of different methods on self reenactment task in NeRSemble dataset~\cite{zhao2023havatar}. From left to right: NeRFBlendShape~\cite{gao2022reconstructing}, NeRFace~\cite{gafni2021dynamic}, HAvatar~\cite{zhao2023havatar} and Ours. Our method can reconstruct details like beards, teeth, eyes, etc. with high quality.}
  \label{fig:self reenactment}
\end{figure*}
\begin{figure*}
  \centering
  \includegraphics[width=1.0\linewidth]{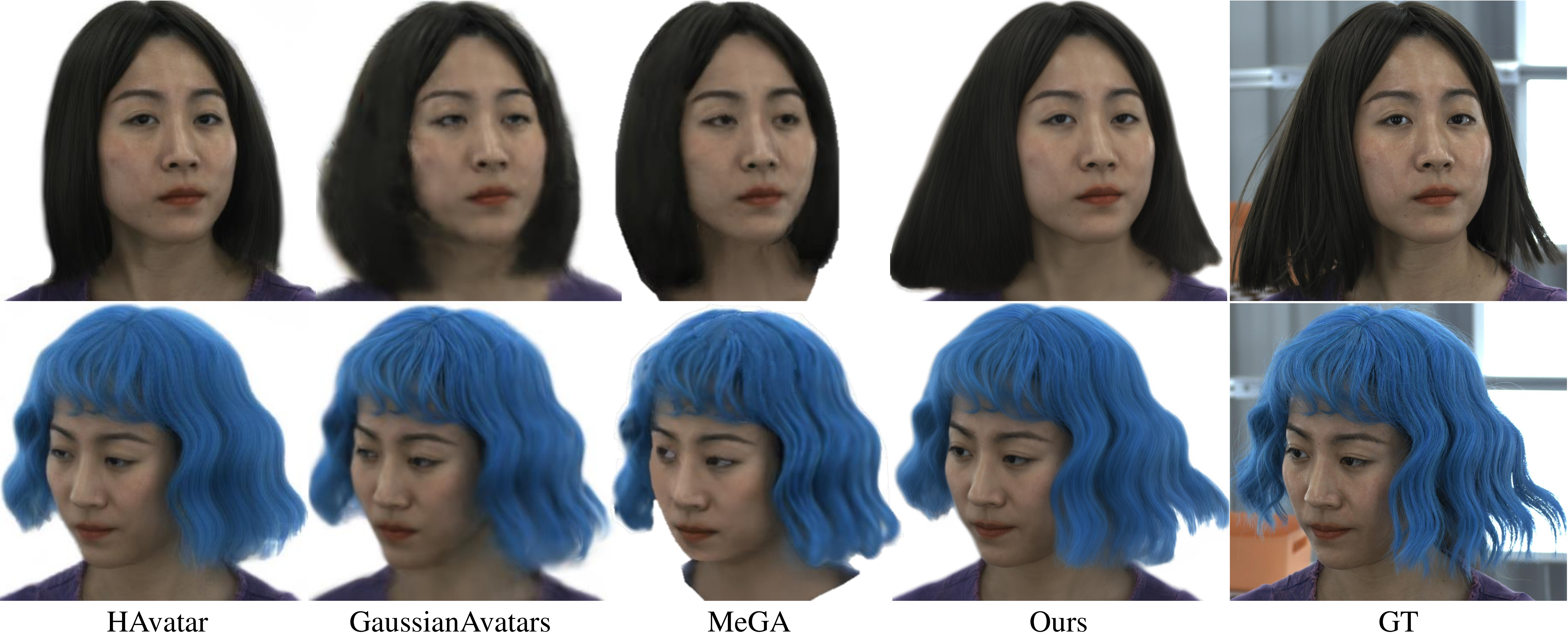}
  \caption{Qualitative comparisons of different methods on self reenactment task with dynamic hairs in the self-captured dataset. From left to right: HAvatar~\cite{zhao2023havatar}, GaussianAvatars~\cite{qian2024gaussianavatars}, MeGA~\cite{wang2024mega} and Ours. Our method can reconstruct details with high quality.}
  \label{fig:self reenactment with hairs}
\end{figure*}
\subsection{Implementation Details}

In the experiment, we use 15 sets of data, with 10 from NeRSemble~\cite{kirschstein2023nersemble}, 2 from multi-view video data from HAvatar~\cite{zhao2023havatar}, and 3 extra self-captured and constructed data containing hair motion as there is no complete non-rigid dynamic hair trajectory (e.g., nodding and swinging at different speeds) in NeRSemble~\cite{kirschstein2023nersemble}.
For the 10 identities from NeRSemble, each set contains 2500 to 3000 frames, 16 cameras are distributed about 120 degrees in front, and simultaneously capture 2K resolution video. For each identity, We use the sequences marked with "FREE" as the evaluation data, and the rest as the training data. 
For the 2 identities from HAvatar, each set contains 3000 frames, 8 cameras are distributed about 120 degrees in front, and 4K resolution videos are collected simultaneously. Later, we crop the face area and resize all images to 2K resolution.
For the 3 identities from self-captured data, each set contains about 1000 frames, 4 cameras are distributed about 90 degrees in front, and 4K resolution videos are collected simultaneously. Later, we crop the face area and resize all images to 2K resolution.

For data preprocessing, we first remove the background~\cite{Lin2021real}, segment the hair from the head by the face-parsing~\cite{yu2021bisenet}, and extract 68 2D facial landmarks~\cite{bulat2017how} for all the images. Then, for each frame, we use multi-view images to estimate the corresponding 3D landmarks, the expression coefficients, and the head pose by fitting the Basel Face Model (BFM)~\cite{gerig2018morphable} to the extracted 2D landmarks. Note that we define the 3D landmarks as the usual 68 landmarks with vertices indexed as multiples of 100 in the BFM vertices.

\subsection{Training Details}
During the geometry-guided initialization stage, we use an Adam optimizer, and set the learning rate to $1 \times 10^{-3}$ for all the networks and $1 \times 10^{-4}$ for the neutral 3D landmarks $\boldsymbol{P}_0$. 
Then, we train the model for 10000 iterations with a batch size of 4.
During the Gaussian model training stage, we also use an Adam optimizer, and set the learning rate to $1 \times 10^{-4}$ for the three color MLPs $\{\boldsymbol{f}_{col}^{exp}, \boldsymbol{f}_{col}^{pose}, \boldsymbol{f}_{col}^{hair}\}$, the three deformation MLPs $\{\boldsymbol{f}_{def}^{exp}, \boldsymbol{f}_{def}^{pose}, \boldsymbol{f}_{def}^{hair}\}$, and the three attribute MLPs $\{\boldsymbol{f}_{att}^{exp}, \boldsymbol{f}_{att}^{pose}, \boldsymbol{f}_{att}^{hair}\}$, $1 \times 10^{-5}$ for the neutral positions $\boldsymbol{X}_0$ and the point-wise feature vectors $\boldsymbol{F}_0$, $1 \times 10^{-4}$ for the neutral rotation $\boldsymbol{Q}_0$, $3 \times 10^{-4}$ for the neutral scale $\boldsymbol{S}_0$, $1 \times 10^{-3}$ for the neutral opacity $\boldsymbol{Q}_0$ and $1 \times 10^{-4}$ for the super resolution network $\boldsymbol{\Psi}$. Finally, we train the Gaussian model for 600000 iterations with a batch size of 1 until fully convergence.

\begin{table*}
\centering
\setlength{\tabcolsep}{13pt}
\begin{tabular}{c|c|c|c|c|c}
\hline
Method             & PSNR $\uparrow$    & SSIM $\uparrow$    & LPIPS (512) $\downarrow$   & LPIPS (2K) $\downarrow$   & FID (2K) $\downarrow$   \\
\hline
\hline
NeRFBlendShape     & 25.91              & 0.836              & 0.123                      & 0.229                     & 54.80              \\
NeRFace            & 27.14              & 0.849              & 0.147                      & 0.234                     & 65.11              \\
HAvatar            & 27.19              & 0.883              & 0.064                      & 0.209                     & 31.06              \\
Ours (w/o SR)      & \textbf{27.82}     & \textbf{0.887}     & 0.080                      & 0.202                     & 45.50              \\
Ours               & 27.70              & 0.883              & \textbf{0.056}             & \textbf{0.098}            & \textbf{18.50}     \\
\hline
\end{tabular} 
\caption{Quantitative evaluation results of NeRFBlendShape~\cite{gao2022reconstructing}, NeRFace~\cite{gafni2021dynamic}, HAvatar~\cite{zhao2023havatar}, our method without super resolution and our full method on self reenactment task in NeRSemble dataset~\cite{zhao2023havatar} and HAvatar dataset~\cite{zhao2023havatar}.}
\label{tab:evaluation}
\end{table*}

\begin{table}[t]
\centering
\begin{tabular}{c|c|c|c|c}
\hline
Method             & PSNR $\uparrow$    & SSIM $\uparrow$    & LPIPS $\downarrow$   & FID $\downarrow$       \\
\hline
\hline
HAvatar            & 26.25              & 0.879              & 0.063                & 36.34        \\
GaussianAvatars    & 24.21              & 0.823              & 0.181                & 65.27        \\
MeGA               & 25.13              & 0.880              & 0.165                 & 55.34       \\
Ours               & \textbf{27.05}     & \textbf{0.883}     & \textbf{0.060}       & \textbf{30.53}         \\
\hline
\end{tabular} 
\caption{Quantitative evaluation results of the other SOTA methods and our method on self reenactment task with the motion of the hair in our self-captured video.}
\label{tab:3d_consistency_hair}
\end{table}

\begin{figure*}
  \centering
  \includegraphics[width=1.0\linewidth]{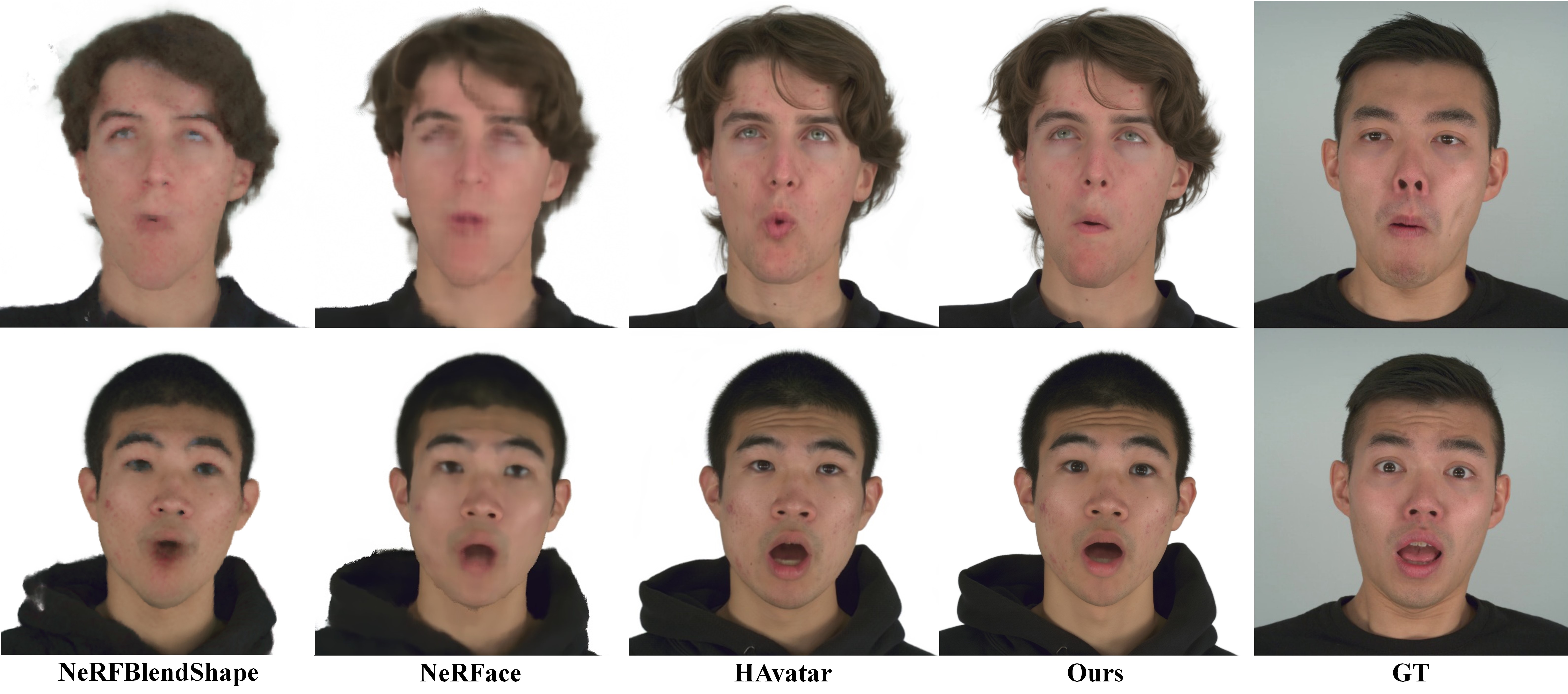}
  \caption{Qualitative comparisons of different methods on cross-identity reenactment task. From left to right: NeRFBlendShape~\cite{gao2022reconstructing}, NeRFace~\cite{gafni2021dynamic}, HAvatar~\cite{zhao2023havatar} and Ours. Our method synthesizes high-fidelity images while ensuring the accuracy of expression transfer.}
  \label{fig:cross-identity reenactment}
\end{figure*}
\begin{figure*}
  \centering
  \includegraphics[width=1.0\linewidth]{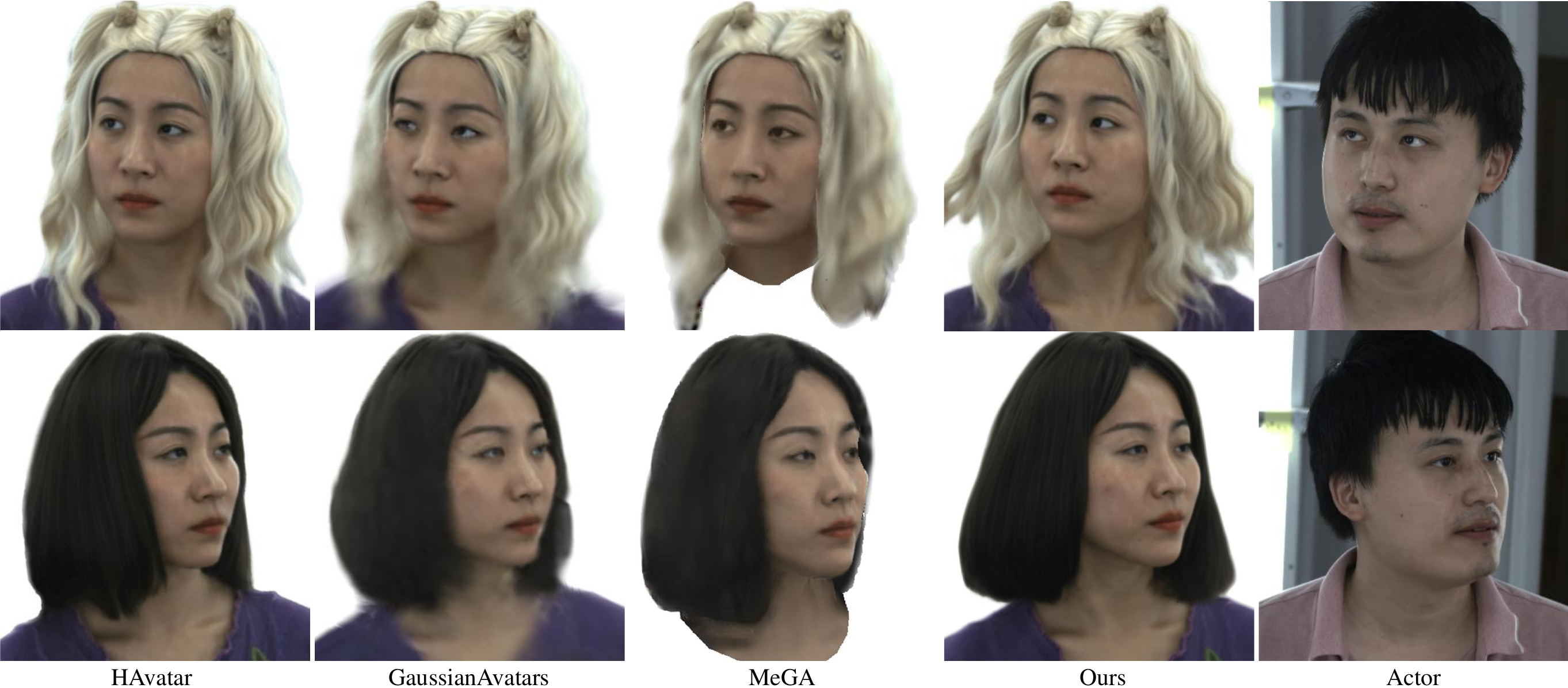}
  \caption{Qualitative comparisons of different methods on cross-identity reenactment task with dynamic hairs in the self-captured dataset. From left to right: HAvatar~\cite{zhao2023havatar}, GaussianAvatars~\cite{qian2024gaussianavatars}, MeGA~\cite{wang2024mega} and Ours. Our method synthesizes high-fidelity images while ensuring the accuracy of hair movement.}
  \label{fig:cross-identity reenactment with hairs}
\end{figure*}
\begin{table}[ht]
\centering
\begin{tabular}{c|c|c|c}
\hline
Method             & PSNR $\uparrow$    & SSIM $\uparrow$    & LPIPS $\downarrow$          \\
\hline
\hline
NeRFBlendShape     & 25.43              & 0.812              & 0.148                        \\
NeRFace            & 26.65              & 0.825              & 0.151                        \\
HAvatar            & 27.13              & 0.880              & 0.65                        \\
Ours               & \textbf{27.58}     & \textbf{0.882}     & \textbf{0.059}                \\
\hline
\end{tabular} 
\caption{Quantitative evaluation results of the other SOTA methods and our method on 3D consistency.}
\label{tab:3d_consistency}
\end{table}

\begin{figure*}
  \centering
  \includegraphics[width=1.0\linewidth]{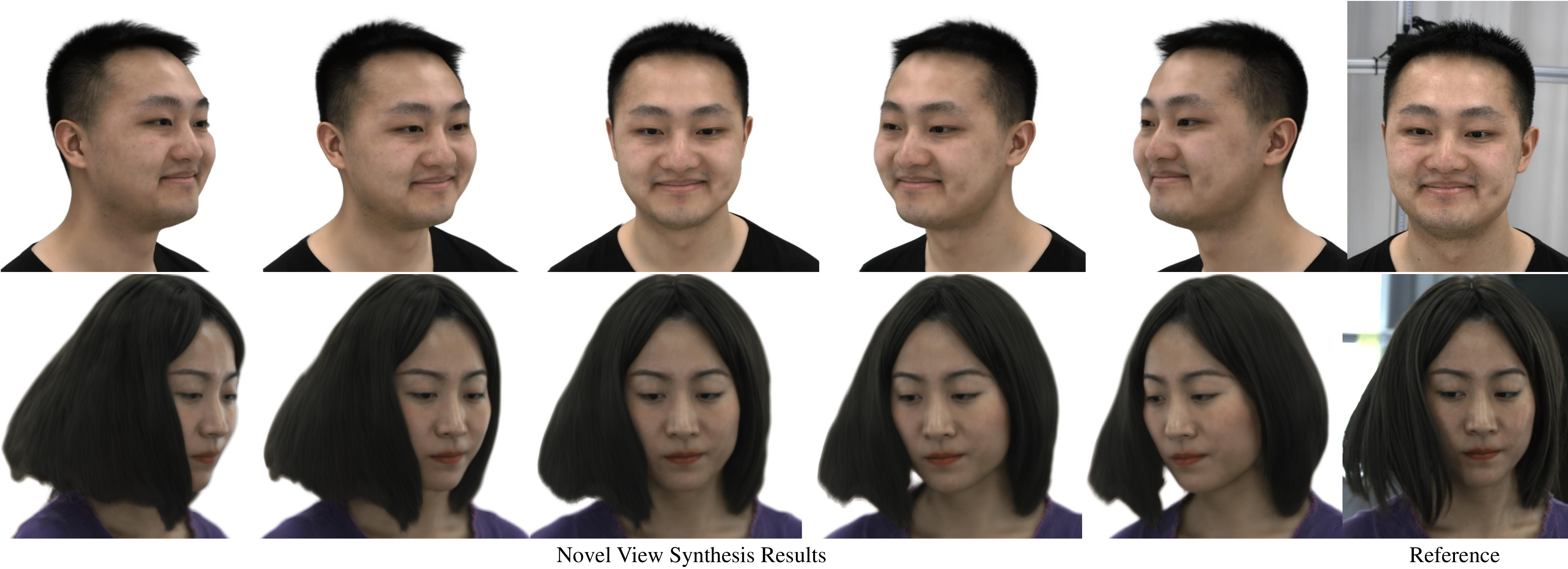}
  \caption{Novel view synthesis results of our method. Top: we use 8-view synchronized videos for training the avatar. Bottom: we use 4-view synchronized videos for training the avatar with dynamic hais.}
  \label{fig:novel_view}
\end{figure*}


\subsection{Results and Comparisons}

\textbf{Self Reenactment.} In this section, we first compare our method with existing SOTA methods in qualitative experiments on self reenactment task. Specifically, NeRFace~\cite{gafni2021dynamic} uses a deep MLP to fit an expression condtioned dynamic NeRF. The current SOTA method HAvatar~\cite{zhao2023havatar} introduces 3DMM template prior and uses a deep convolutional network to generate a human head NeRF represented by three planes from a mesh template with expression. Note, HAvatar leverages the GAN framework using the adversarial loss function to force the network to generate details that are not view-consistent. For a fair comparison, we remove this part and use VGG perceptual loss as Sec.~\ref{subsec:avatar_learning} instead. 

Qualitative results on self reenactment task are shown in the Fig.~\ref{fig:self reenactment}. Our method can accurately reconstruct pixel-level high-frequency details such as beards, teeth, and hair. Besides, our method can achieve expression transfer more accurately, such as eye movements in the figure.

Next, we conduct a quantitative evaluation for the four methods on 5 identities and 6 cameras using the evaluation split. The evaluation metrics include: Peak Signal-to-Noise Ratio (PSNR), Structure Similarity Index (SSIM), Learned Perceptual Image Patch Similarity (LPIPS)~\cite{zhang2018the} and Fréchet Inception Distance (FID)~\cite{martin2017gans}. Note, we calculate FID by comparing the distribution of all the training images and all the rendered images. As the task mainly focuses on the reconstruction of the head, we use face-parsing~\cite{yu2021bisenet} to remove the body parts in the image to eliminate their impact in the experiment. 
As shown in Tab.~\ref{tab:evaluation}, our method demonstrates a slight improvement in PSNR and SSIM compared with previous methods, and a significant improvement in LPIPS and FID, which means that our method can generate more high-frequency details.
Note that HHAvatar only made additional changes to the position and attributes of the Gaussians in the hair part, and the final rendering method remains the same as the previous version\cite{xu2023gaussianheadavatar}. 
Therefore, when the dataset is organized chronologically and the hair maintains relatively static compared to the head, the reconstructed results are consistent with the previous version\cite{xu2023gaussianheadavatar}. 
When the dataset is not in chronological order, simply canceling the change in hair inertia will result in the same reconstruction as the previous version\cite{xu2023gaussianheadavatar}.

We also compare our method with existing SOTA methods in qualitative experiments on self reenactment task with dynamic hairs.
Qualitative results on self reenactment task are shown in the Fig.~\ref{fig:self reenactment with hairs}. 
It can be seen that all other methods fail to capture the accurate physical motion of the hair component and the hair part is blurry.
Furthermore, as a result of the occlusion created by hair during motion, the face itself appears blurry.
Because all existing methods neglect the kinematic attributes of hair, none of them can successfully complete self reenactment task with the motion of the hair.
By modeling hair separately from the human head and introducing the kinematic features of the hair, our method can accurately reconstruct high-frequency details of the hair. Besides, our method can capture the realistic physical motion of hair.

Quantitative results on self reenactment task with dynamic hairs are shown in Tab.~\ref{tab:3d_consistency_hair}. 
As shown in Tab.~\ref{tab:3d_consistency_hair}, our method demonstrates a slight improvement in PSNR and SSIM compared with previous methods, and a significant improvement in LPIPS, which means that our method can generate more high-frequency details.

\textbf{Cross-Identity Reenactment.} We qualitatively compare our method with the above SOTA methods on cross-identity reenactment task. As shown in Fig.~\ref{fig:cross-identity reenactment}, our method is able to synthesize higher-fidelity images with more accurate expression transfer and richer emotions.

We also qualitatively compare our method with the above SOTA methods on cross-identity reenactment task with dynamic hairs. As shown in Fig.~\ref{fig:cross-identity reenactment with hairs}, other methods did not consider the status of the hair at the previous moment, resulting in no reasonable physical motion of the hair during driving.
Therefore, those methods fail to capture the accurate physical motion of the hair component and the hair part is blurry.
Our method is able to synthesize higher-fidelity images with more accurate hair motion.

\textbf{Novel View Synthesis.} In this section, we show the results of novel view synthesis as shown in the top of Fig.~\ref{fig:novel_view} shown. In this case, we use video data from 8 views for training and render the image at a new viewpoint.
Next, we quantitatively evaluate the 3D consistency of our method and compare it with other SOTA methods mentioned above. Specifically, we select the 5 identities as above and use the video data from 8 cameras for training while the other 8 hold-out cameras for evaluation. The evaluation metrics include PSNR, SSIM and LPIPS (512 resolution). The results are shown in Tab.~\ref{tab:3d_consistency}. Our method outperforms other methods in 3D consistency.

We also show the results of novel view synthesis with dynamic hairs, and the results are shown in the bottom of Fig.~\ref{fig:novel_view}. 
In this case, we use our self-captured video data from 4 views for training and render the image at five new viewpoints. 
As shown in the bottom of Fig.~\ref{fig:novel_view}, our method can maintain 3D consistency and synthesize high-fidelity images with accurate hair motion.

\subsection{Ablation Study}
\begin{figure}
  \centering
  \includegraphics[width=1.0\linewidth]{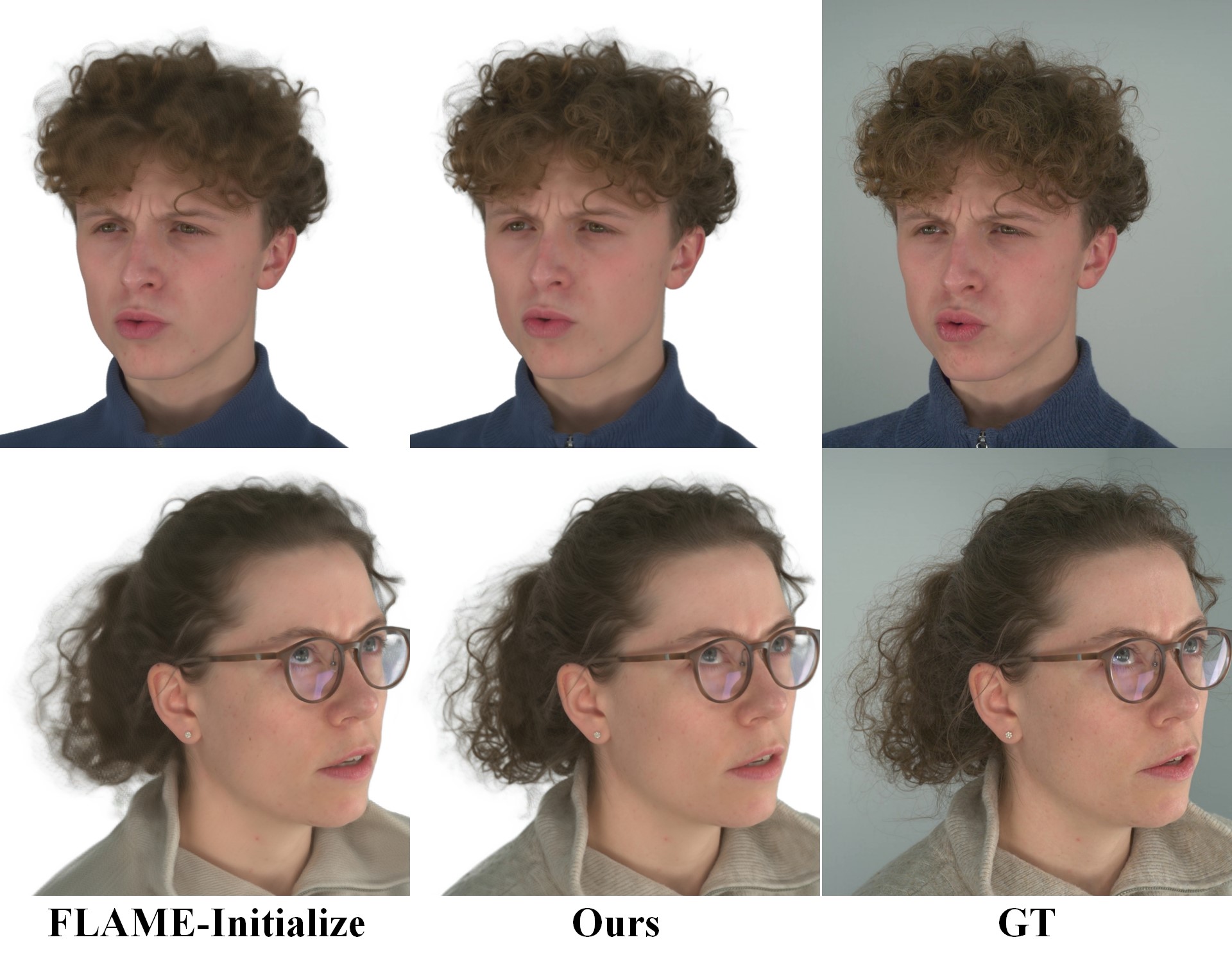}
  \caption{Ablation study on the initialization strategies: FLAME-initialization and our geometry-guided initialization. Our strategy ensures the hair strands away from the head are well reconstructed.}
  \label{fig:ablation_initialization}
\end{figure}

\begin{table}[ht]
\centering
\begin{tabular}{c|c|c|c}
\hline
Method             & PSNR $\uparrow$    & SSIM $\uparrow$    & LPIPS $\downarrow$          \\
\hline
\hline
FLAME-Init         & 28.73              & 0.875              & 0.123                        \\
Mesh-Deform        & 28.83              & 0.874              & 0.116                        \\
Ours               & \textbf{28.94}     & \textbf{0.876}     & \textbf{0.108}                \\
\hline
\end{tabular} 
\caption{Quantitative evaluation results of the other two ablation baselines and ours on self reenactment task.}
\label{tab:ablation}
\end{table}

\noindent\textbf{Ablation on Initialization Strategies.} In order to verify the effectiveness of our geometry-guided initialization strategy~\ref{subsec:initialization}, we compare it with the strategy to use the FLAME model for initialization (FLAME-Init). Specifically, after fitting a FLAME model through multi-view data, we first subdivide the FLAME mesh 4 times and use the neutral vertices as the positions of the neutral Gaussians. Then, the expression deformation MLP is optimized to learn the displacement of FLAME vertices. We set the per-vertex feature to zeros, while randomly initialize the parameters of the expression color MLP. The initialization of other variables is the same as our strategy.
Qualitative results are shown in the Fig.~\ref{fig:ablation_initialization}. Due to the lack of initialization for the hair and shoulders in FLAME-Init, the points to model these parts are offset from nearby vertices, which leads to sparseness of the Gaussians, resulting in blurring.

\begin{figure}[ht]
  \centering
  \includegraphics[width=1.0\linewidth]{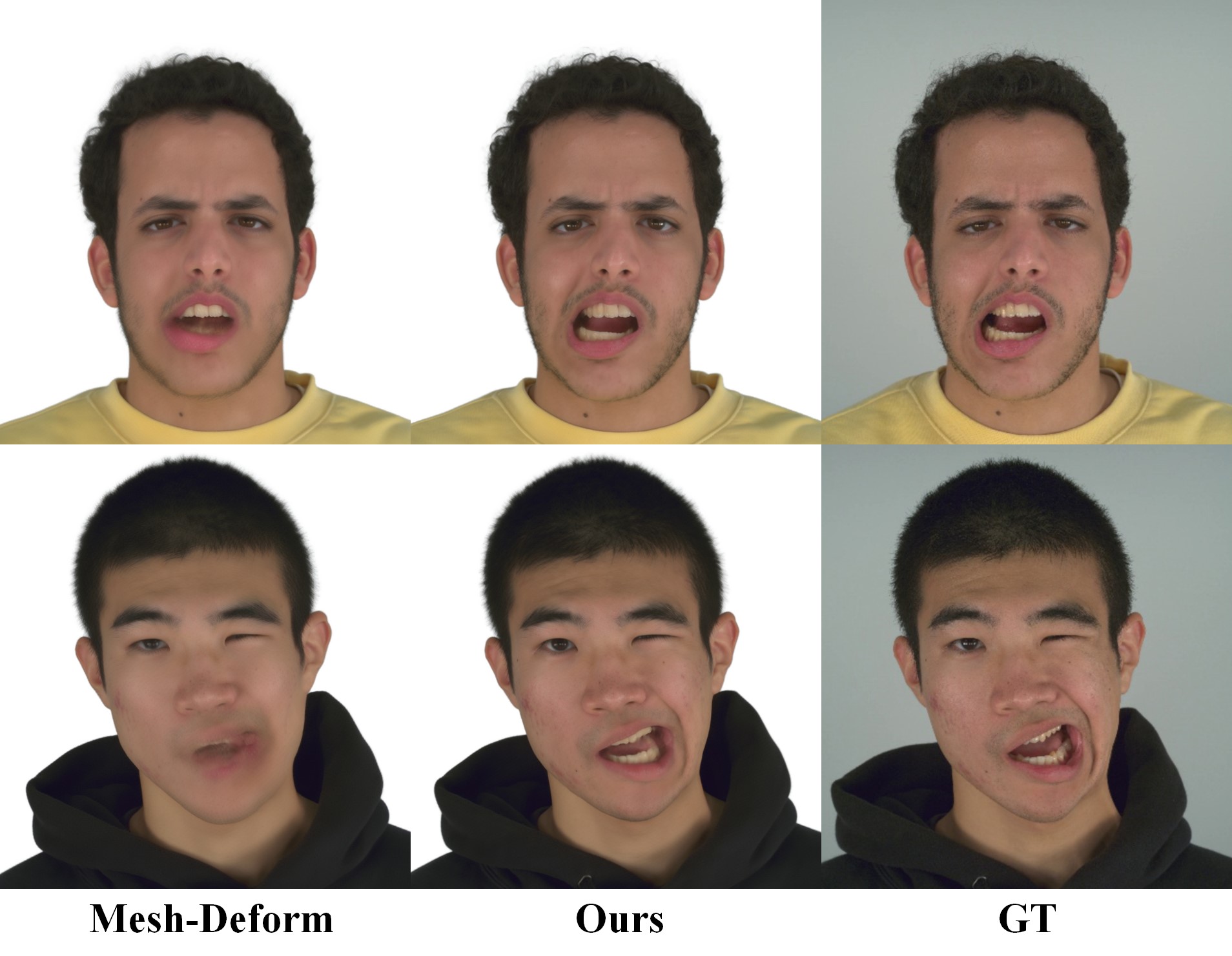}
  \caption{Ablation study on the deformation modeling: mesh LBS-based deformation and our fully learned deformation. Our approach can learn complex and exaggerated expressions.}
  \label{fig:ablation_deformation }
\end{figure}

\noindent\textbf{Ablation on Deformation Modeling Approaches.} We compare our fully learned deformation field with the previous mesh-based deformation (Mesh-Deform). Specifically, we migrate the method in INSTA~\cite{zielonka2022instant} for controlling the NeRF deformation to our Gaussians. First we fit a 3DMM mesh template. Then, for each Gaussian point, find the closest face on the mesh, and calculate the deformation gradient to estimate the displacement. 
Qualitative results are shown in the Fig.~\ref{fig:ablation_deformation }. For some expressions that cannot be captured well by the 3DMM mesh template, our method can learn accurate deformation, thereby achieving the modeling of complex expressions.
Quantitative results are shown in the Fig.~\ref{tab:ablation}. Our method outperforms both the two ablation baselines on PSNR, SSIM and LPIPS metrics.

\begin{figure}[!t]
  \centering
  \includegraphics[width=1.0\linewidth]{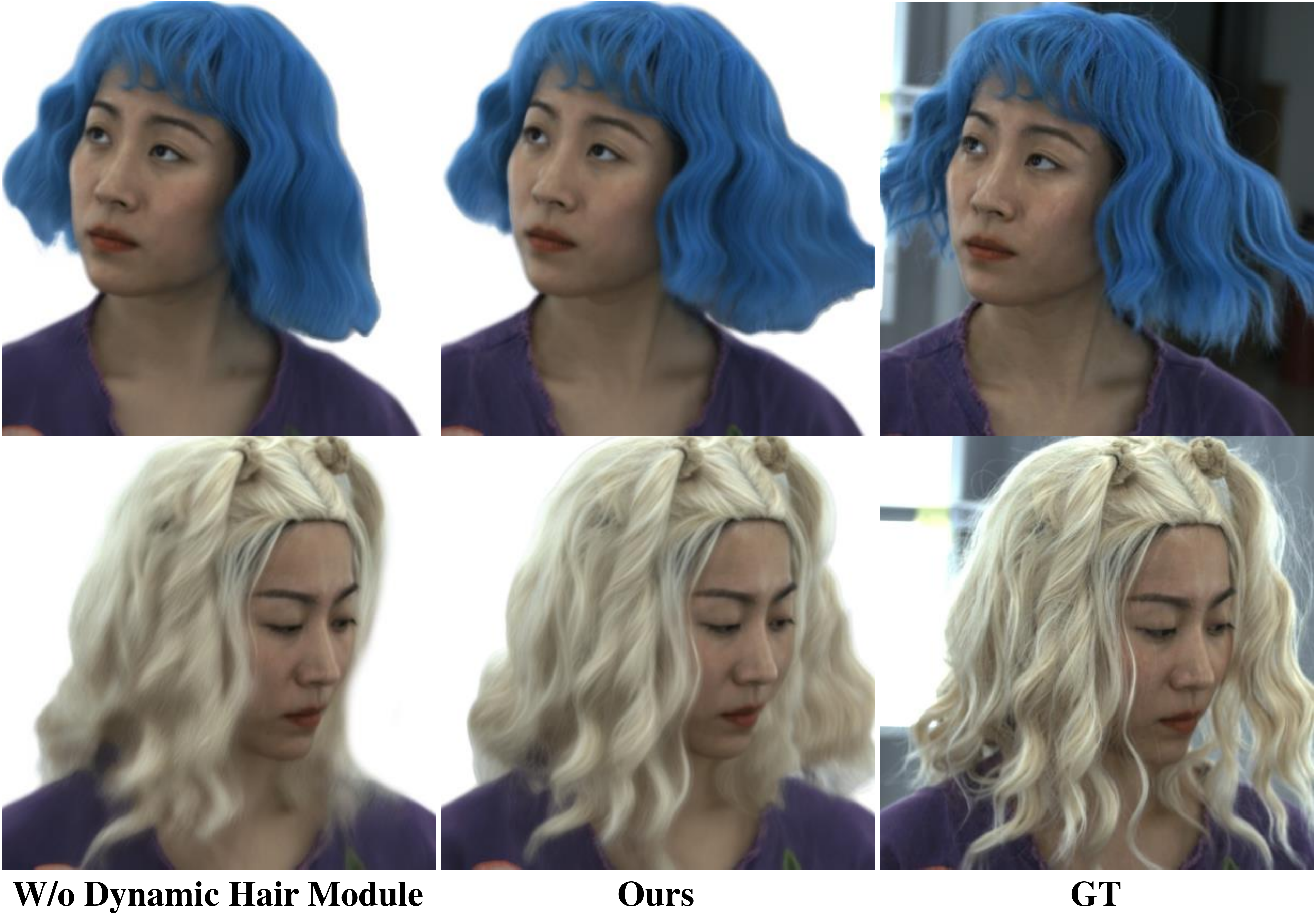}
  \caption{Ablation study on the dynamic hair module: our approach can capture the motion of the hair and reconstruct more accurate hair.}
  \label{fig:ablation_deformation_hair}
\end{figure} 
\begin{table}[!t]
\centering
\begin{tabular}{c|c|c|c}
\hline
Method             & PSNR $\uparrow$    & SSIM $\uparrow$    & LPIPS $\downarrow$          \\
\hline
\hline
W/o Dynamic Hair         & 26.23              & 0.825              & 0.123                        \\
W/o Occlusion Perception        & 26.83              & 0.874              & 0.116                        \\
Full Model               & \textbf{27.02}     & \textbf{0.876}     & \textbf{0.108}                \\
\hline
\end{tabular} 
\caption{Quantitative evaluation results of the baseline and our full model on self reenactment task.}
\label{tab:ablation_hair}
\end{table}
\begin{figure}[!t]
  \centering
  \includegraphics[width=1.0\linewidth]{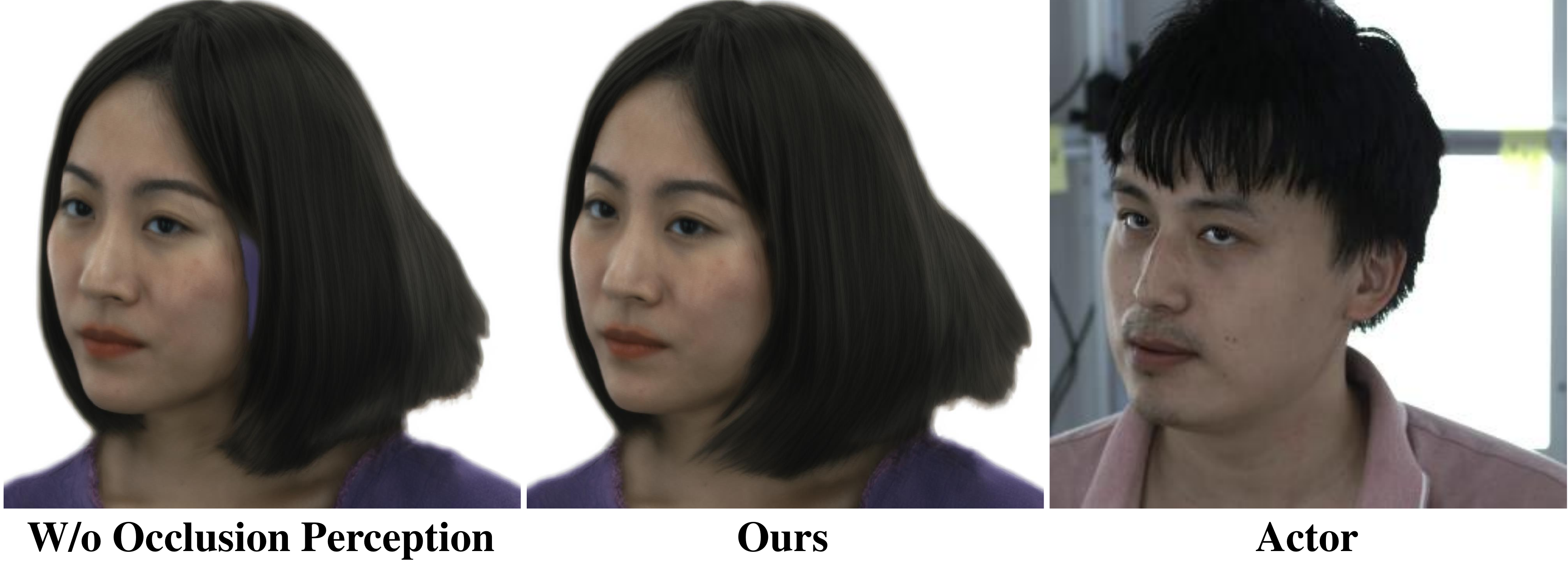}
  \caption{Ablation study on the dynamic hair module: our approach can capture the motion of the hair and reconstruct more accurate hair.}
  \label{fig:ablation_occ}
\end{figure} 

\noindent\textbf{Ablation on Hair Dynamic Module.} To assess the effectiveness of independently modeling hair, we designed an experiment in which we kept modeling the head and the hair in the same manner.
Qualitative results are shown in the Fig.~\ref{fig:ablation_deformation_hair}.
It can be seen that if the hair is not modeled separately from the head, not only will the hair not have realistic physical motion, but it will also produce blurring.
Therefore, it is necessary to model hair separately.
Quantitative results are shown in the Tab.~\ref{tab:ablation_hair}. 
It can be seen that our method outperforms both the two ablation baselines on PSNR, SSIM and LPIPS metrics.

\noindent\textbf{Ablation on Occlusion Perception Module.} To assess the effectiveness of the mask that considers occlusion, we designed an experiment in which we used alpha rendering~\footnote{https://github.com/ashawkey/diff-gaussian-rasterization} for separating the hair and the head.
Qualitative results are shown in the Fig.~\ref{fig:ablation_occ}.
It can be seen that due to the lack of consideration for occlusion, the hole may sometimes occur in the face during driving because of the hair motion.
Therefore, it is necessary to consider the occlusion.
Quantitative results are shown in the Tab.~\ref{tab:ablation_hair}. 
It can be seen that our method outperforms both the two ablation baselines on PSNR, SSIM and LPIPS metrics.

\begin{figure}[!t]
  \centering
  \includegraphics[width=0.9\linewidth]{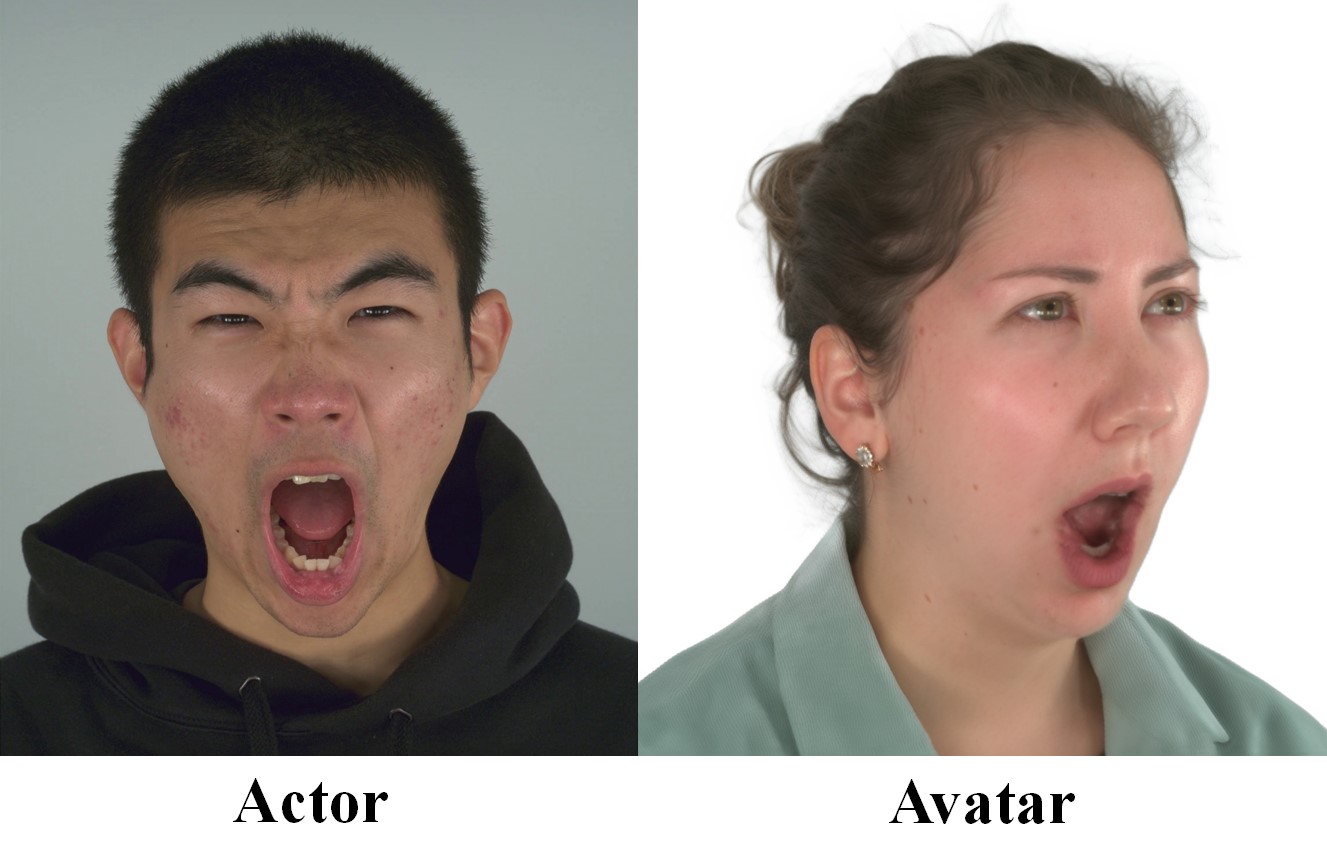}
  \caption{Failure case: our method produce relatively less exaggerated results.}
  \label{fig:failure2}
\end{figure}
\section{Discussion and Conclusion}

\noindent\textbf{Ethical Considerations.} Our method is capable of creating artificial portrait videos, which have the potential to disseminate misinformation, influence public perceptions, and erode confidence in media sources. 

\noindent\textbf{Limitation.} 
For the tongue and teeth inside the mouth, blurring is sometimes produced in our method due to the lack of tracking methods.
On the other hand, the reconstructed head avatar cannot make expressions other than those in the training set. Therefore, when the actor's expression is too exaggerated, our method will output relatively less exaggerated results as shown in Fig.~\ref{fig:failure2}.
When the head movement speed in the training set is excessively high, leading to complete hair coverage over facial features, the fitting of 3DMM expression coefficients will fail. 
Hence, it is essential to avoid moving the head too rapidly.

\noindent\textbf{Conclusion.} In this paper, we propose HHAvatar, a novel representation for head avatar reconstruction, which leverages dynamic 3D Gaussians controlled by a fully learned expression deformation and models the physical motion of the hairs. 
Experiments demonstrate our method can synthesize ultra high-fidelity images while modeling exaggerated expressions and capturing the realistic physical motion of the hair.
In addition, we propose a well-designed minute-level initialization strategy to ensure the training convergence and a temporal module to fuse time series information.
We believe our HHAvatar will become the mainstream direction for head avatar reconstruction in the future.


\bibliographystyle{IEEEtran}
\bibliography{main}

\end{document}